\pdfoutput=1

\documentclass[11pt]{article}

\usepackage{acl}

\usepackage{times}
\usepackage{latexsym}

\usepackage[T1]{fontenc}

\usepackage[utf8]{inputenc}

\usepackage{microtype}

\usepackage{inconsolata}

\usepackage{graphicx}
\usepackage{amsmath,array}
\usepackage{amsthm}
\usepackage{threeparttable}
\usepackage{multirow}
\usepackage{amsfonts}
\usepackage{caption}
\usepackage{bbold}
\usepackage{placeins}
\usepackage{color}
\usepackage{mathtools}
\usepackage{makecell}
\usepackage{booktabs}

\usepackage{enumitem}

\title{\textsc{Araida}: Analogical Reasoning-Augmented Interactive Data Annotation}

\author{
Chen Huang$^{\spadesuit\clubsuit}$, \quad
Yiping Jin$^{\heartsuit}$, \quad
Ilija Ilievski$^{\diamondsuit}$, \quad
Wenqiang Lei$^{\spadesuit\clubsuit}$\thanks{Correspondence to Wenqiang Lei.}, \quad
Jiancheng Lv$^{\spadesuit\clubsuit}$
\\
$^{\spadesuit}$College of Computer Science, Sichuan University, China \\
$^{\clubsuit}$Engineering Research Center of Machine Learning and Industry Intelligence, \\Ministry of Education, China\\
$^{\heartsuit}$NLP Group, Pompeu Fabra University, Spain\\
$^{\diamondsuit}$ISEM, National University of Singapore, Singapore \\
\texttt{huangc.scu@gmail.com, yiping.jin@upf.edu, wenqianglei@gmail.com}
}

\begin{document}
\maketitle
\begin{abstract}
Human annotation is a time-consuming task that requires a significant amount of effort. To address this issue, interactive data annotation utilizes an annotation model to provide suggestions for humans to approve or correct. However, annotation models trained with limited labeled data are prone to generating incorrect suggestions, leading to extra human correction effort. To tackle this challenge, we propose \textsc{Araida}, an analogical reasoning-based approach that enhances automatic annotation accuracy in the interactive data annotation setting and reduces the need for human corrections. \textsc{Araida} involves an error-aware integration strategy that dynamically coordinates an annotation model and a k-nearest neighbors (KNN) model, giving more importance to KNN's predictions when predictions from the annotation model are deemed inaccurate. Empirical studies demonstrate that \textsc{Araida} is adaptable to different annotation tasks and models. On average, it reduces human correction labor by 11.02\% compared to vanilla interactive data annotation methods.
\end{abstract}

\section{Introduction}
Data annotation is a challenging task that involves a tradeoff between annotation quality and budget. While some platforms offer a cost-effective solution by relying on ML models to annotate data automatically~\footnote{For example, \url{https://aws.amazon.com/sagemaker/groundtruth/}.}, the quality of such annotations is often compromised~\cite{wang2022whose}. It is particularly true in the \textbf{limited data annotation} scenario where the annotation budget is limited or when unlabeled data are scarce~\cite{ringger2007active,chaudhary2021reducing, huang2024selective}.

\begin{figure}
    \centering
    \includegraphics[width=0.46\textwidth]{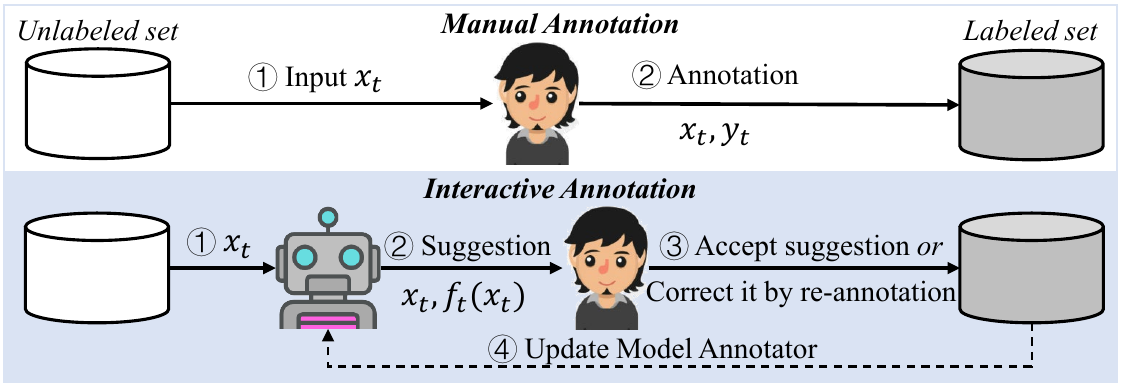}
    \caption{Comparison between manual annotation and interactive annotation.}
    \label{fig:enter-labelddd}
\end{figure}

Human-machine \textbf{interactive annotation} methods were introduced to reduce annotation effort while maintaining annotation quality~\cite{klie-etal-2018-inception,klie-etal-2020-zero,le2021interactive}. As illustrated in Fig.~\ref{fig:enter-labelddd}, these methods introduce an \textit{annotation model} to suggest labels (\textit{model annotations}) to human annotators. The annotators accept a suggested label if it is correct. Otherwise, they have to correct the label manually. Compared to manual annotation, interactive annotation requires less human effort because human annotators only have to verify the model annotations instead of coming up with an answer from scratch, leading to potential speedup of the annotation process~\citep{klie-etal-2020-zero}.

\begin{figure*}[t]
\centering
  \includegraphics[width=1\textwidth]{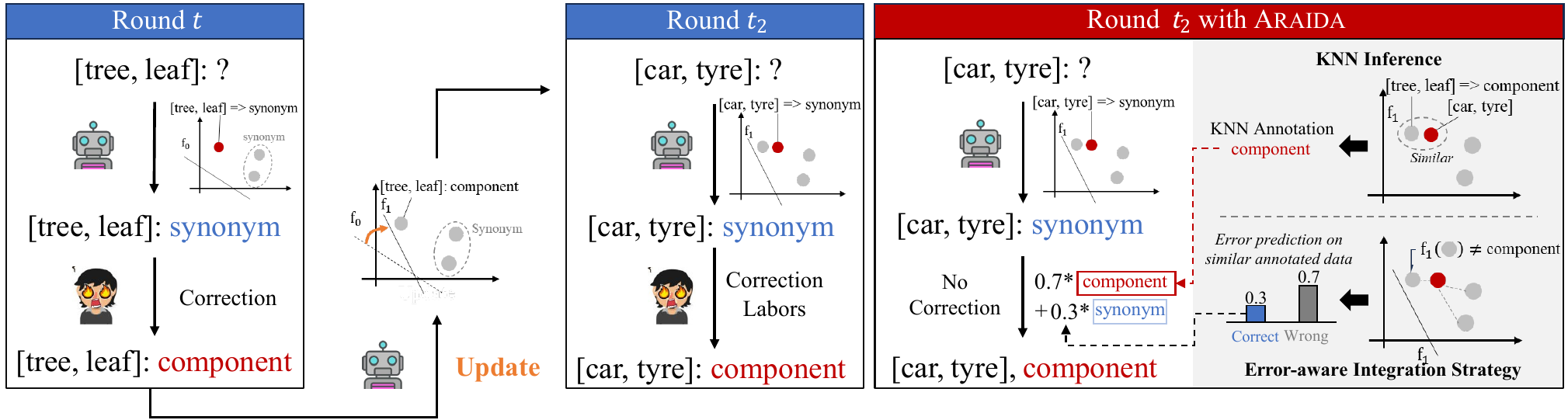}
    \caption{
      Example on span relation annotation. An under-trained annotation model results in more suggestion errors and increases human correction effort. \textcolor{red}{\textsc{Araida}} improves the model annotation accuracy via the KNN model and the error-aware integration strategy for dynamical coordination of annotations.
    } 
    \label{fig:03-03}
\end{figure*}

Evidently, the annotation model's accuracy is crucial because incorrect suggestions require additional human effort to rectify. Existing methods update the annotation model based on previously accepted or corrected data (\textit{ground-truth annotation}), aiming to reduce human corrections by improving prediction accuracy at each iteration \cite{klie-etal-2020-zero, WU2022364}.
However, in the context of limited data annotation, the annotation model lacks sufficient labeled training data to reach a reasonable accuracy and is prone to providing incorrect suggestions~\cite{rietz2021cody}. For example, in the span relation annotation example shown in Fig.~\ref{fig:03-03} (\textcolor{blue}{blue}), the annotation model continues to make mistakes on similar examples ([\textit{car}, \textit{tyre}]) even after the human annotator corrects the label `[\textit{tree}, \textit{leaf}]$=>$\textit{component}'. As a result, this leads to more human corrections. Such a problem is crucial for interactive annotation and has been identified by recent work \cite{rietz2021cody}, but it has yet to be addressed.


Inspired by cognitive studies on efficient learning~\cite{lake2017building,lake2015human,mitchell2021abstraction}, finding that the human brain can learn from a few examples because our brain is continuously building analogies during the learning process of concepts to facilitate comprehension, we propose \textbf{A}nalogical \textbf{R}easoning-\textbf{A}ugmented \textbf{I}nteractive \textbf{D}ata \textbf{A}nnotation (\textsc{Araida}), which is designed to improve interactive annotation under the limited data annotation setting. \textsc{Araida} provides an annotation reference to the annotation model by retrieving previously human-labeled examples in the proximity of the example in consideration using the k-nearest neighbors (KNN) method.
As illustrated in Fig.~\ref{fig:03-03}(\textcolor{red}{red}), the final suggestion combines the model annotation and the annotation reference provided by KNN via an error-aware integration strategy. This strategy dynamically coordinates the annotation model and KNN, giving more importance to KNN's prediction if the predicted label from the annotation model is estimated to be inaccurate.  


We conduct simulated experiments for the limited data annotation task and estimate the human annotation effort based on the number of human corrections (or the number of suggestion errors) following \citet{hwa2000sample} and \citet{kristjansson2004interactive}.
We test \textsc{Araida} on different word-level and sentence-level annotation tasks, combining with different annotation models (i.e., classic and LLM-based models). The result shows that \textsc{Araida} consistently improves different annotation models' accuracy across various tasks. On average, it reduces human corrections by 11.02\%. 
Further analysis attributes this improvement to the few-shot capability of the KNN module and the error-aware integration strategy that effectively synergizes complementary annotations. In summary, our contributions are as follows:

\vspace{-\topsep}
\begin{itemize}[leftmargin=*]
    \setlength\itemsep{-0.3em}
    \item Calling attention to the limited data annotation scenario. We highlight the under-trained problem of the annotation model, which is crucial in practice but overlooked in interactive annotation.
    \item Introducing \textsc{Araida} that involves a KNN module and an error-aware integration strategy to alleviate the under-trained problem by facilitating coordination between the two model annotators (i.e., the vanilla annotation model and KNN).
    \item Demonstrating the efficacy of \textsc{Araida} in enhancing suggestion accuracy, reducing human corrections, and showcasing its flexibility to combine with various annotation models through extensive experiments.
\end{itemize}
\vspace{-\topsep}

\section{Related Work}
Our research is tied to interactive data annotation, human analogical reasoning (KNN), and retrieval-based language models. We provide a literature review and highlight our differences.

\paragraph{Interactive Data Annotation.} Interactive data annotation aims to reduce human annotation effort by incorporating an annotation model that suggests labels to human annotators during an interactive process~\citep{klie-etal-2018-inception,klie-etal-2020-zero,le2021interactive}. 
The annotation model must be sample-efficient because, when we start a new annotation task, there are few labeled examples to learn from. Current studies focus on employing active learning~\citep{klie-etal-2018-inception,laws-etal-2011-active,casanova2020reinforced, li2021fitannotator, huang-etal-2023-reduce} to prioritize annotating examples more likely to improve model accuracy. While active learning can reduce the required training data to some extent, it may not be effective in limited data annotation scenarios or when complex hypotheses or semantics are to be learned~\cite{dasgupta2005coarse, rietz2021cody}. 
Another approach is to employ LLMs for automatic data annotation, which have demonstrated strong performance under zero-shot and few-shot settings~\cite{he2023annollm, gilardi2023chatgpt}. However, such performance might not be consistent for difficult tasks, as they may even perform worse than fine-tuned small language models \cite{xiao-etal-2023-freeal}. Regardless of whether we use active learning and whether we use a classic or LLM-based annotation model, our empirical evidence demonstrates that \textsc{Araida} can effectively decrease the amount of human corrections required.

\paragraph{KNN and Analogical Reasoning.} While KNN has been extensively utilized in NLP community~\citep{wang2019simpleshot, liu2023knn, wang2022k}, its underlying mechanism is often overlooked. To shed light on this, cognitive studies~\citep{lake2017building,lake2015human,mitchell2021abstraction} revealed that the KNN inference process aligns with human analogical reasoning, enabling efficient learning \citep{lake2017building, lake2015human}.
In particular, analogical reasoning establishes connections between relevant aspects of the current task and past experiences, forming abstractions that enhance human reasoning capabilities~\citep{mitchell2021abstraction}. In this context, KNN facilitates sample-efficient learning by leveraging similarities between the example to be labeled and previously annotated examples, resulting in exemplary solutions~\citep{NIPS2016_65fc52ed}, which reduce the training data requirement.

\paragraph{Retrieval-Based Language Models.} There is growing interest in enhancing the output of language models by incorporating a retrieval module (usually KNN or alike) that interpolates with a datastore built from the training data~\cite{khandelwal2019generalization, kassner-schutze-2020-bert}. Compared to vanilla language models, retrieval-based models ground the predictions in labeled training examples, potentially yielding better explainability and sample efficiency~\citep{asai-etal-2023-retrieval}.
This approach has shown promising results in tasks such as machine translation \cite{khandelwal2020nearest, liu2023knn}, named entity recognition \cite{wang2022k}, and question answering \cite{kassner-schutze-2020-bert}. While some studies have explored the use of dynamically adjusted combination weights between the language model and the retrieval module~\citep{wang2021non,zheng2021adaptive,jiang-etal-2021-learning}, our method differs significantly for two main reasons: 1) Different tasks. We are the pioneers in introducing KNN to the interactive data annotation task, whereas these methods are primarily designed for machine translation. 2) Different techniques. We adjust the weight by estimating the error of model predictions for each data point (e.g., sentence), whereas these methods learn the weight for each token without error estimation.

\section{\textsc{Araida}: The Proposed Method}
We present \textsc{Araida}, an analogical reasoning-based method for interactive data annotation that provides an annotation reference to the annotation model by retrieving previously human-labeled examples in the proximity of the example in consideration. We detail the KNN inference module in Section~\ref{knn} and the error-aware integration strategy in Section~\ref{error}. Finally, the optimization details are provided in Section \ref{opei}.

\paragraph{Task Formalization and Overview.} Let $X$ denote the dataset that needs to be annotated, with $C$ being the number of classes. 
Given a data batch $x_t$ at time $t$, the annotation model $f_t$ predicts label vectors $f_t(x_t) \in R^{|x_t|\times C}$, and the KNN module $g_t$ infers label vectors $g_t(x_t) \in R^{|x_t|\times C}$ using previously annotated data stored in a datastore $A_t$. 
Then, we estimate the probability $\lambda_t \in R^{|x_t|\times 1}$ of the annotation model's predictions $f_t(x_t)$ being reliable, i.e., $argmax(f_t(x_t)) = y_t$, where $argmax(\cdot)$ returns indices of the classes with the highest predicted probability and $y_t\in R^{|x_t|\times 1}$ are the ground truth labels. 
Finally, we use $\lambda_t$ to weigh the two predictions $f_t(x_t)$ and $g_t(x_t)$ through a linear weighted combination:
\begin{equation}
\label{main_equ}
    F_t(x_t) = \lambda_t \cdot f_t(x_t) + (1-\lambda_t) \cdot g_t(x_t).
\end{equation}

Notably, closed-source language models such as ChatGPT produce discrete labels rather than predicted distributions. Therefore, we cannot combine its predictions with KNN's using linear combination. In such case, we use binary values ($0$ or $1$) for $\lambda_t$, which acts as a function allocation to determine whether $f_t$ or $g_t$ should apply to each example.
Once the human approves or corrects the final suggestions, the datastore $A_t$ is updated with the newly arrived data batch and its corresponding labels. In addition, the annotation model $f_t$ (if applicable), KNN module $g_t$, and weighting strategy $\lambda_t$ are updated via back-propagation.

\subsection{KNN Inference}
\label{knn}

We utilize a weighted KNN to perform inference, defined as $g_t(x_t^i) = \frac{\sum_{a \in \rho_i} w_a y_a}{\sum_{a \in \rho_i} w_a}$, where $\rho_i \in A_t$ is the $k$ nearest neighbors of each example $x_t^i$, $y_a$ corresponds to the human annotation of each neighbor $a \in \rho_i$. 
The similarity between $x_t^i$ and $a$ is measured by $w_a=\frac{1}{d(x_t^i, a)}$, where $d(x_t^i, a)=\|w_{knn}(x_t^i-a)\|_2$ is a distance metric parameterized by $w_{knn}$. We use the similarity measure to retrieve $\rho_i$. To avoid overconfidence in KNN inference, we apply label smoothing to the labels of the retrieved neighbors. Specifically, we set $y_a = y_a (1-\alpha) + \alpha/C$, where $\alpha=1-\frac{1}{C}$.

\paragraph{Datastore Maintenance Strategy.} The datastore $A_t$ consisting of historically annotated data grows in size as the interactive annotation continues, causing the KNN's retrieval to be less time-efficient. To address this issue, we impose a constraint on the maximum datastore size using a pre-defined hyperparameter. We propose a class-aware maintenance strategy. Precisely, if $A_t$ exceeds its budget, data that is from the majority class~\footnote{The majority class refers to the class with the highest frequency in $A_t$.} and most similar to its class prototype\footnote{The class prototypes are the average of the feature vectors in $A_t$ that belong to each class.} is discarded first. This strategy ensures that the datastore contains as many labeled data from different classes as possible while minimizing the impact on the class prototype. Appendix~\ref{dyenicx} and \ref{bufferthis} present experiments using different datastore sizes and maintenance strategies.

\subsection{Error-aware Integration Strategy}
\label{error}
\paragraph{Motivation.} A popular method to combine annotations from two models (i.e., annotation model and KNN) is to use a weighted linear combination with a constant weight~\footnote{Equivalent to when $\lambda_t(x_t)$ in Eq.\ref{main_equ} is a constant.}~\citep{liu2023knn, wang2022k}. However, assuming that one model consistently outperforms the other on all unlabeled data is unrealistic. Furthermore, both models are updated with each new batch of data in interactive data annotation, and their relative performance will alter, making it infeasible to find the optimal weight through a one-off hyperparameter tuning.
To address this issue, we propose an error-aware integration strategy that automatically assigns weights to different models, relying more on KNN inference when the annotation model's prediction is estimated to be inaccurate.

\paragraph{Error Estimation of Model Annotation.} We base on the intuition that if the model $f_t$ consistently makes mistakes on previous examples similar to the current data point $x_t^i$, then its prediction $f_t(x_t^i)$ will likely be incorrect. To achieve this, we parameterize the integration strategy $\lambda_t$ as a neural network to learn from the customized input. 

\vspace{-\topsep}
\begin{itemize}[leftmargin=*]
    \setlength\itemsep{-0.3em}
    \item \underline{Customized Inputs}. 
    For each data point $x_t^i$, we derive the input $\mathbb{x}_t^i$ to the integration strategy $\lambda_t$, which considers the local error estimation $E_t^i$ and local density $D_t^i$. 
    Specifically, $E_t^i$ is a vector with elements $e_{t,j}^i=\mathbb{1}[argmax(f_t(a_j)) = y_{a_j}]$ indicates if the annotation model $f_t$ predicted correctly on each annotated example $a_j\in \rho_i$ in the $k$ nearest neighbors of $x_t^i$.
    The local density $D_t^i$ is a distance vector, with each element being $d(x_t^i, a_j)$. 
    These two vectors are combined using the element-wise multiplication operator $\odot$ to create the input: $\mathbb{x}_t^i=D_t^i \odot E_t^i - D_t^i \odot (1-E_t^i)$. Notably, $\mathbb{x}_t^i$ measures the error regularity of $x_t^i$ among its neighborhood, as the more positive values in the vector $\mathbb{x}_t^i$, the less likely $f_t$ would make an error on $x_t^i$. 
    \item \underline{Learning Objectives}. We collect the ground truth labels $y_t$ through human feedback. To optimize the error-aware integration strategy $\lambda_t$, we use a mean squared error (MSE) loss, denoted as $\ell_d^t(y_t, f_t(x_t), \lambda_t) = \text{MSE}(\mathbb{1}[argmax(f_t(x_t))=y_t], \lambda_t(\mathbb{x}_t))$, where $\mathbb{1}[\cdot]$ indicates whether the ground truth labels $y_t$ are the same as the predictions $f_t(x_t)$. The purpose of this loss function is to guide $\lambda_t$ by encouraging it to predict errors made by $f_t$.
\end{itemize}
\vspace{-\topsep}



\subsection{Optimization of \textsc{Araida}}
\label{opei}
To simplify the optimization process, we independently optimize the annotation model $f_t$, KNN model $g_t$, and error-aware integration strategy $\lambda_t$. We treat human feedback $y_t$ as the ground truth following previous studies on interactive annotation \cite{klie-etal-2018-inception,klie-etal-2020-zero,le2021interactive}.
It is worth noting that \textsc{Araida} supports any task-specific annotation model and uses its corresponding loss function $\ell_f$ to update the parameters. Combining the $\ell_d$ loss and the negative log-likelihood loss $\ell_g$ to optimize KNN, we formulate the final loss function as follows:
\begin{equation}
\small
\begin{split}
    \mathcal{L}(f,g,\lambda)
    =\sum_{i=1}^{B_t}& \ell_f(y_i, f_t(x_i)) + \ell_g(y_i, g_t(x_i)) \\ & + \ell_d(y_i, f_t(x_i), \lambda_t),
\end{split}
\end{equation}
where $B_t$ represents the total data accumulated until round $t$.
There are two challenges to optimizing this objective function. Firstly, the operator used in KNN to retrieve the $k$ nearest neighbors is not differentiable. To address this problem, we utilize the Gumbel-softmax-based re-parameterization trick~\citep{jang2016categorical} to facilitate the optimization process. Secondly, the loss function $\mathcal{L}$ presents a bi-level optimization problem, where the optimization of $\lambda_t$ is nested within the optimization problems of $f_t$ and $g_t$. As a result, we update $f_t$, $g_t$, and $\lambda_t$ iteratively using a coordinate-descent approach. Formally, at each optimization iteration $k$, we have network parameters $\theta_f^k$, $\theta_g^k$, and $\theta_\lambda^k$ corresponding to $f^k$, $g^k$, and $\lambda^k$. The update procedures are as follows:
\begin{equation}
\begin{split}
    \theta_f^{k+1} &= \theta_f^k - \bigtriangledown_f \mathcal{L}(f, g^k, \lambda^k), \\
    \theta_g^{k+1} &= \theta_g^k - \bigtriangledown_g \mathcal{L}(f^{k}, g, \lambda^k), \\
    \theta_\lambda^{k+1} &= \theta_\lambda^k - \bigtriangledown_\lambda \mathcal{L}(f^{k+1}, g^{k+1}, \lambda).
\end{split}
\end{equation}

\section{Experiments}
We conduct extensive experiments to assess \textsc{Araida}'s effectiveness in the limited data annotation scenario. Our primary focus is to assess whether \textsc{Araida} can decrease the human effort required for corrections by providing more precise annotations at various stages of the annotation process (see Section \ref{sec2}). Furthermore, we perform a comprehensive examination to investigate the behavior and impact of KNN and the error-aware integration strategy (see Section \ref{sec3}). Additional analysis of our error-aware integration strategy is presented in Section \ref{why}. Lastly, we analyze the sensitivity of the parameters in Appendices~\ref{sec4}.

\begin{table}[]
\centering
\resizebox{0.485\textwidth}{!}{
\begin{tabular}{p{1.2cm}rp{5.2cm}}
\hline
\textbf{Dataset} &  \textbf{\# Val.} & \textbf{Classes}\\ \hline 
WN18RR & 3,034 & Hypernym; Derivation; Member; Component; Synset; Synonym; Verb group; Instance of hypernym;\\\hline
FreeBase & 5,116 & Contains; Country; Track\_role; Profession; Group\_role; Adjoins; Film\_release; Nutrient\\\hline
IMDB & 5,000 & Positive; Negative\\\hline
SST-5 & 1,101 & Strong positive; Positive; Neutral; Negative; Strong negative\\\hline
\end{tabular}%
}
\caption{Statistics of datasets. For each dataset, we randomly sample 5K examples from the original training dataset to form the unlabeled data, and the validation dataset is taken from the original dataset. Table \ref{tab:yes} presents the mapping from the original categories to the categories we use.}
\label{tab:data}
\end{table}

\begin{table*}[]
\centering
\resizebox{0.99\textwidth}{!}{%
\begin{tabular}{l|llll|llll}
\toprule
\multicolumn{1}{c|}{\multirow{3}{*}{\textbf{\begin{tabular}[c]{@{}c@{}}Annotation \\ Model\end{tabular}}}} & \multicolumn{4}{c|}{\textbf{Without Active Learning (AL)}} & \multicolumn{4}{c}{\textbf{With Active Learning (AL)}} \\ \cline{2-9} 
\multicolumn{1}{c|}{} & \multicolumn{2}{c|}{\textbf{Word-level Annotation}} & \multicolumn{2}{c|}{\textbf{Sentence-level Annotation}} & \multicolumn{2}{c|}{\textbf{Word-level Annotation}} & \multicolumn{2}{c}{\textbf{Sentence-level Annotation}} \\ \cline{2-9} 
\multicolumn{1}{c|}{} & \multicolumn{1}{c|}{WN18RR} & \multicolumn{1}{c|}{FreeBase} & \multicolumn{1}{c|}{IMDB} & \multicolumn{1}{c|}{SST-5} & \multicolumn{1}{c|}{WN18RR} & \multicolumn{1}{c|}{FreeBase} & \multicolumn{1}{c|}{IMDB} & \multicolumn{1}{c}{SST-5} \\ \midrule
Dist./FT & 50.44±1.02 & 32.47±12.37 & 70.18±8.43 & 36.02±3.14 & 47.77±0.91 & 26.92±10.85 & 66.98±6.31 & 35.20±3.76 \\
Dist./FT + \textsc{Araida}& \textbf{52.16±1.37} & \textbf{43.02±6.43} & \textbf{79.33±2.81} & \textbf{37.21±3.03} & \textbf{49.54±0.84} & \textbf{39.06±4.57} & \textbf{75.84±1.14} & \textbf{37.02±2.50} \\ \hline
LLaMa2 & 31.35±1.89 & 24.41±1.77 & 80.21±2.64 & 37.83±1.64 & 31.35±1.89 & 24.41±1.77 & 80.21±2.64 & 37.83±1.64 \\
LLaMa2 + \textsc{Araida}& \textbf{45.15±1.96} & \textbf{38.20±1.91} & \textbf{88.47±2.03} & \textbf{42.03±1.88} & \textbf{46.33±2.01} & \textbf{38.79±1.96} & \textbf{89.68±2.25} & \textbf{42.61±1.91} \\ \hline
LLaMa2$_{sft}$ & 58.24±2.79 & 53.11±1.38 & 94.06±9.02 & 46.84±6.30 & 59.28±2.57 & 55.39±2.01 & 95.13±10.15 & 47.45±6.17 \\
LLaMa2$_{sft}$ + \textsc{Araida} & \textbf{60.74±2.33} & \textbf{55.23±1.46} & \textbf{95.15±9.32} & \textbf{49.62±5.98} & \textbf{61.02±2.18} & \textbf{56.71±2.09} & \textbf{95.88±12.27} & \textbf{49.51±6.03}  \\ \bottomrule
\end{tabular}%
}
\caption{Machine cumulative accuracy (MCA) scores using various methods. We report the averaged results across varying amounts of data. LLaMa2 with AL has identical performance as LLaMa2 because the vanilla LLaMa2 model is not updated during the interactive annotation process. Thus, the data order does not impact the performance. When combined with \textsc{Araida}, the performances w/ and w/o AL are different because the KNN and integration strategy models are updated.}
\label{main_results1}
\end{table*}

\subsection{Experimental Setup}
\paragraph{Tasks \& Datasets.} We experiment with word- and sentence-level annotation tasks. These tasks have been highlighted as crucial in various web applications ~\cite{yao2021interactive, marcos2020information, lee2022promptiverse}. 
To simulate the scenario of limited data annotation, we follow \citet{dou2019investigating} by imposing dataset size restrictions, ranging from $1K$ to $5K$. Table \ref{tab:data} overviews the dataset statistics.

\vspace{-\topsep}
\begin{itemize}[leftmargin=*]
    \item \textbf{Word-level annotation}. We focus on the knowledge graph completion task, which annotates the semantic classes of input word pairs (e.g., `[\textit{tree}, \textit{leaf}]$=>$\textit{component}'). We use two benchmark knowledge graph datasets in our experiments, namely the WN18RR~\citep{dettmers2018convolutional}\footnote{\url{https://paperswithcode.com/dataset/wn18rr}} and Freebase~\citep{bollacker2008freebase}\footnote{\url{https://www.microsoft.com/en-us/download/details.aspx?id=52312}} dataset. We experiment with the eight most frequent classes for each dataset.
    \item \textbf{Sentence-level annotation}. We consider the sentiment classification task and experiment on two benchmark datasets, including SST-5~\citep{socher2013recursive}\footnote{\url{https://nlp.stanford.edu/sentiment/code.html}}, and IMDB~\citep{maas-EtAl:2011:ACL-HLT2011}\footnote{\url{https://www.kaggle.com/datasets/lakshmi25npathi/imdb-dataset-of-50k-movie-reviews}}. SST-5 dataset contains categories on a scale of 1-5 while IMDB contains two categories (positive/negative).
\end{itemize}

\paragraph{Evaluation Metric.} We aim to minimize the total human corrections (i.e., the total model suggestion errors) annotating a given amount of data using the interactive annotation process. Therefore, we report the \textit{Machine Cumulative Accuracy} (MCA), defined as the total correct suggestions divided by the total suggestions for different dataset sizes. To assess the performance of each method in the limited data annotation scenario, we present the mean and the corresponding standard deviation by varying the dataset size (\{1K, 2K, 3K, 4K, 5K\}).

\paragraph{Annotation Models.} To verify the generalizability of \textsc{Araida}, we apply it in conjunction with different annotation models:
\vspace{-\topsep}
\begin{itemize}[leftmargin=*]
    \item \underline{Classic annotation models}. We utilize lightweight annotation models following previous works~\cite{pmlr-v133-desmond21a, chen2020jit2r, hedderich2021anea}. Specifically, for word-level tasks, we use a distributional model~\cite{roller2014inclusive, kober2021data} with pretrained GloVe word embeddings embedding~\cite{pennington2014glove}\footnote{We did not use contextualized embeddings because the word-level task in our experiment has no context.}. For sentence-level tasks, we use FastText \cite{joulin2017bag} to derive the sentence embeddings. We denote this baseline as \textit{Dist./FT}.
    \item \underline{LLM-based annotators}. We also use large language models (LLMs) as annotation models, whose few-shot and in-context learning capabilities might help with the limited data annotation process. We experiment with LLaMa2-7B~\cite{touvron2023llama} and ChatGPT~\cite{ouyang2022training}~\footnote{The \texttt{gpt-3.5-turbo} checkpoint in OpenAI's API (\url{https://platform.openai.com/docs/models/gpt-3-5}).}. 
    We use zero-shot and few-shot prompts for ChatGPT (denoted as \textit{ChatGPT\textit{$_{zero}$}} and \textit{ChatGPT\textit{$_{few}$}}). Detailed prompts can be found in Table~\ref{tab:prompt}.
    For LLaMa2, we consider both the vanilla \textit{LLaMa2} that uses the same zero-shot prompts as ChatGPT\textit{$_{zero}$} and \textit{LLaMa2\textit{$_{sft}$}}, which is fine-tuned using an open-source toolkit\footnote{\url{https://github.com/Alpha-VLLM/LLaMA2-Accessory}} during the interactive annotation process. Fine-tuning data examples can be found in Table~\ref{tab:llama}.
    
\end{itemize}

\paragraph{Impact of Active Learning.} Regardless of the annotation model, the sequence of the data to annotate also affects the amount of human corrections. Intuitively, if we show unambiguous examples first, few corrections are needed. However, the annotation model and KNN may not learn to handle more challenging examples and may subsequently make more mistakes. Previous studies focused on applying active learning in interactive annotation to enhance the annotation models' sample efficiency~\cite{laws-etal-2011-active, klie-etal-2018-inception, li2021fitannotator}. To study the impact of active learning in the limited data annotation scenario, we compare an uncertainty-based active learning method with random data ordering for different annotation models with and without \textsc{Araida}. Note that we omit the ChatGPT with AL results because we are unable to estimate its prediction uncertainty accurately.

\paragraph{Implementation Details.} All experiments are carried out on a machine with Intel(R) Xeon(R) Gold 5317 CPU @ 3.00GHz and a GeForce RTX 3090 GPU. For simplicity, we implement our integration strategy using a three-layer, fully-connected network with ReLu activation and dropout. For KNN, we set $k=20$ for $\rho_t$. KNN runs in the embedding space of \textit{text-embedding-ada-002}~\cite{neelakantan2022text} when combining with ChatGPT. For Dist./FT and LLaMa2 models, KNN runs in the corresponding model's embedding space. Moreover, we leave the details of LLM prompts and mode fine-tuning examples in Appendix \ref{apendix:llm}.

\begin{figure}
    \centering
\includegraphics[width=0.46\textwidth]{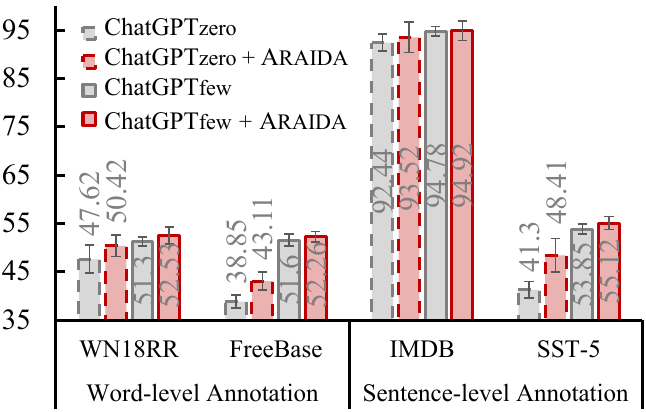}
    \caption{MCA scores using ChatGPT-based methods. We omit the ChatGPT with AL results because we are unable to estimate its prediction uncertainty. \textsc{Araida} can further improve ChatGPT's performance}
    \label{main_results2}
\end{figure}

\subsection{Main Result}
\label{sec2}
We evaluate the effectiveness of \textsc{Araida} in enhancing the model annotation quality, hence reducing human correction effort. Table~\ref{main_results1} and Figure \ref{main_results2} show the machine cumulative accuracy scores averaged across varying amounts of data ($\{1K, 2K, 3K, 4K, 5K\}$) for different experimental settings. We make the following observations:

\textbf{\textsc{Araida} reduces human corrections consistently}. As illustrated in Table \ref{main_results1} and Figure \ref{main_results2}, \textsc{Araida} consistently improves the model suggestion accuracy across different annotation models and annotation tasks. Specifically, it achieves a 13.06\% performance gain in MCA on Dist./FT, 3.85\% on LLaMa2$_{sft}$, 16.80\% on Dist./FT with AL, 2.61\% on LLaMa2$_{sft}$ with AL, 30.48\% on LLaMa2, 8.81\% on ChatGPT$_{zero}$, and 1.55\% on ChatGPT$_{few}$. 
On average, it achieves an 11.02\% performance gain in model annotation accuracy, translating into a reduction in human correction effort in practice. 

\textbf{Active learning does not always help.} When comparing the annotation models with and without active learning, we observe that active learning does not always improve model performance and can even harm the performance in some instances (e.g., Dist./FT). We hypothesize that the more challenging cases selected by active learning might require more training data for the models to correct their predictions~\cite{dasgupta2005coarse, rietz2021cody}, which are unavailable in limited data annotation scenarios. Instead of relying on an annotation model alone, \textsc{Araida} acts as a posthoc ``plug-in'' that fixes the annotation model's mistakes using retrieved annotation references and yields a robust improvement under various settings with and without active learning.

\textbf{LLMs are strong annotation models. \textsc{Araida} can improve them further.} 
LLMs perform better than classic distributional models, especially for sentence-level tasks. Furthermore, LLaMa2 with fine-tuning consistently outperforms the vanilla LLaMa2 model, and few-shot ChatGPT consistently beats its zero-shot counterpart. Interestingly, comparing these two pairs of annotation models, we observe that \textsc{Araida} brings a more substantial improvement to weaker models. When the annotation models are already strong (in the case of LLaMa2$_{sft}$ and ChatGPT$_{few}$), \textsc{Araida} is more conservative in making corrections, yielding a smaller but consistent improvement. It demonstrates \textsc{Araida}'s robustness to combine with annotations models with different performances.

\begin{table}[htb]
\centering
\resizebox{0.48\textwidth}{!}{
\begin{tabular}{l|l|l|l|l}
\toprule
\multicolumn{1}{c|}{\multirow{2}{*}{\textbf{\begin{tabular}[c]{@{}c@{}}Annotation\\ Model\end{tabular}}}} & \multicolumn{2}{c|}{\textbf{Word-level Annotation}} & \multicolumn{2}{c}{\textbf{Sentence-level Annotation}} \\ \cline{2-5} 
\multicolumn{1}{c|}{} & \multicolumn{1}{c|}{\textbf{WN18RR}} & \multicolumn{1}{c|}{\textbf{FreeBase}} & \multicolumn{1}{c|}{\textbf{IMDB}} & \multicolumn{1}{c}{\textbf{SST-5}} \\ \hline

\multicolumn{5}{c}{Dist./FT}\\\hline
\textsc{Araida}  & \textbf{52.16$\pm$1.37} & \textbf{43.02$\pm$6.43} & \textbf{79.33$\pm$2.81} & \textbf{37.21$\pm$3.03} \\
- \textit{w/o} KNN & 50.44$\pm$1.02 & 32.47$\pm$12.37 & 70.18$\pm$8.43 & 36.02$\pm$3.14 \\
- \textit{w/o} $f(\cdot)$ & 50.28$\pm$3.01 & 41.52$\pm$5.87 & 78.48$\pm$0.68 & 35.02$\pm$1.37 \\
- \textit{w/} const. & \multicolumn{1}{l|}{51.85$\pm$4.77} & \multicolumn{1}{l|}{41.78$\pm$6.78} & \multicolumn{1}{l|}{78.58$\pm$3.94} & 37.07$\pm$2.68 \\
 \hline

 \multicolumn{5}{c}{LLaMa2$_{sft}$}\\\hline
\textsc{Araida}           & \textbf{60.74$\pm$2.33} & \textbf{55.23$\pm$1.46} & \textbf{95.15$\pm$9.32} & \textbf{49.62$\pm$5.98} \\ 
- \textit{w/o} KNN        & 58.24$\pm$2.79          & 53.11$\pm$1.38          & 94.06$\pm$9.02          & 46.84$\pm$6.30  \\
- \textit{w/o} $f(\cdot)$ & 45.23$\pm$2.58          & 39.82$\pm$3.41          & 89.13$\pm$0.55          & 42.93$\pm$1.71  \\
- \textit{w/} const.      & 58.24$\pm$2.79          & 53.11$\pm$1.38          & 94.06$\pm$9.02          & 46.84$\pm$6.30  \\
 \hline
 
\multicolumn{5}{c}{ChatGPT$_{zero}$}\\\hline
\textsc{Araida}  & \textbf{50.42$\pm$1.37} & \textbf{43.11$\pm$1.83} & \textbf{93.52$\pm$1.24} & \textbf{48.41$\pm$1.06} \\
- \textit{w/o} KNN & 47.62$\pm$2.89 & 38.85$\pm$2.23 & 92.44$\pm$0.95 & 41.30$\pm$1.72 \\
- \textit{w/o} $f(\cdot)$ &  48.73$\pm$2.64 & 41.30$\pm$5.04 & 90.61$\pm$0.26 & 45.84$\pm$1.12 \\
- \textit{w/} const. & 48.73$\pm$2.64 & 41.30$\pm$5.04 & 92.44$\pm$0.95 & 45.84$\pm$1.12 \\ \midrule
\multicolumn{5}{c}{ChatGPT$_{few}$}\\\hline
\textsc{Araida}  & \textbf{52.53$\pm$1.67} & \textbf{52.26$\pm$3.44} & \textbf{94.92$\pm$1.02} & \textbf{55.12$\pm$1.36} \\
- \textit{w/o} KNN & 51.30$\pm$1.72 & 51.60$\pm$3.15 & 94.78$\pm$0.99 & 53.85$\pm$2.03 \\ 
- \textit{w/o} $f(\cdot)$    & 48.73$\pm$2.64 & 41.30$\pm$5.04 & 90.61$\pm$0.26 & 45.84$\pm$1.12 \\
- \textit{w/} const. & 51.30$\pm$1.72 & 51.60$\pm$3.15 & 94.78$\pm$0.99 & 53.85$\pm$2.03 \\ \bottomrule
\end{tabular}
}
\caption{Ablation study on the KNN and the error-aware integration strategy modules. We report the MCA scores using various methods, averaging results with different amounts of data. Error-aware integration strategy effectively coordinates the two annotators.}
\label{main_resultsddd}
\end{table}

\begin{figure*}[htb!]
    \centering
    \includegraphics[width=1\textwidth]{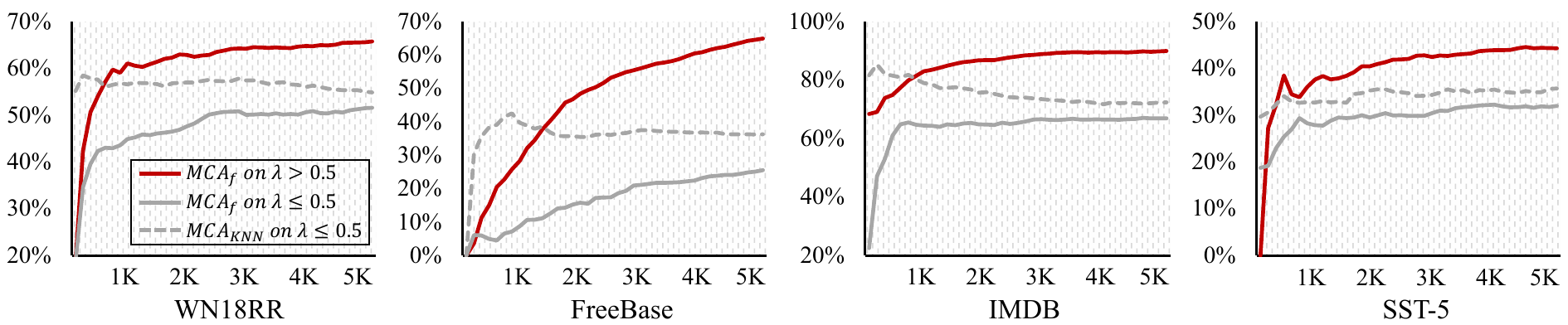}
    \caption{Analyzing our integration strategy with the Dist./FT model. The solid lines show the MAC scores of the annotation model $f(\cdot)$, separated by examples with $\lambda > 0.5$ (higher weights assigned to the annotation model $f(\cdot)$) and $\lambda \leq 0.5$ (higher weights assigned to KNN). The dotted line shows KNN's performance on the latter set.}
    \label{fig:enter-labeldd}
\end{figure*}

\subsection{Ablation Study}
\label{sec3}
This section aims to perform a comprehensive examination to investigate the behavior and impact of the KNN and out integration strategy.
We conduct an ablation study to analyze the effectiveness of each component. In particular, we consider the following baselines:
\vspace{-\topsep}
\begin{itemize}[leftmargin=*]
    \item \textbf{\textsc{Araida} w/o KNN:} Using the annotation model alone to suggest labels. Equivalent to $\lambda_t(\cdot)=1$.
    \item \textbf{\textsc{Araida} w/o f($\cdot$):} Using KNN alone to suggest labels. Equivalent to $\lambda_t(\cdot)=0$.
    \item \textbf{\textsc{Araida} w/ const.:} Using a constant $\lambda_t^*$ value for all examples, in contrast to \textsc{Araida}, which varies $\lambda_t$ for different examples. In our experiments, we report the result with the best $\lambda_t^*$ tuned on the validation set. 
\end{itemize}
The ablation test results are presented in Table~\ref{main_resultsddd}. Due to limited space, we omit the result for vanilla LLaMa2, which is much weaker than other LLM-based baselines. The detailed observations are provided below.

\textbf{KNN is a strong stand-alone annotator.} Table \ref{main_resultsddd} reveals that although KNN (\textit{\textsc{Araida} w/o f($\cdot$)}) does not match the performance of \textsc{Araida}, it obtains comparable accuracy as using the annotation model alone (\textit{\textsc{Araida} w/o KNN}). This result is surprising, given KNN's simplicity compared to the annotation models (including LLMs). Such the results also highlight that significance and effectiveness of analogical reasoning, allowing humans to reason more effectively \cite{mitchell2021abstraction}.

\textbf{Error-aware integration strategy effectively coordinates the two annotators.} Table~\ref{main_resultsddd} shows that \textsc{Araida}'s error-aware integration strategy achieves consistent performance gain compared to using a constant weight $\lambda_t^*$. This in turn also confirms the original intention behind our construction of the error-aware integration strategy: it is infeasible to find the optimal weight through a one-off hyperparameter tuning. Here, it is worth noting that ChatGPT outputs discrete labels rather than probabilistic vectors. In this case, the constant strategy reduces into an indicator function: if $\lambda_t^* > 0.5$, $F_t(x)=f_t(x), \forall x_t$; otherwise, $F_t(x)=g_t(x), \forall x_t$. In this case, \textit{\textsc{Araida} w/ const.} prefers ChatGPT or KNN based on their performance. However, it cannot improve the annotation quality beyond the individual components (i.e., the KNN and the annotation model) due to the discrete output of ChatGPT. 

\subsection{Qualitative Analysis}
\label{why}
To shed light on how the error-aware integration strategy works, we measure the cumulative accuracy of the annotator model and KNN throughout the annotation process for each dataset and present the result with the Dist./FT annotation model in Figure~\ref{fig:enter-labeldd}. We also separate the cases where $\lambda > 0.5$ (higher weights assigned to the annotation model $f(\cdot)$) and $\lambda \leq 0.5$ (higher weights assigned to KNN). Our observations are as follows.

Firstly, we observe that there is a substantial gap between the two solid lines ($MCA_f$ on $\lambda > 0.5$ and $MCA_f$ on $\lambda \leq 0.5$), showing that our error estimation model effectively identifies cases where $f(\cdot)$ is likely to error and assigns it a lower weight. Secondly, KNN reaches a reasonable accuracy much faster than the annotation model at the initial stage of the interactive annotation process. Even as the number of annotated examples increases, its accuracy is still higher than the annotation model $f(\cdot)$ when $\lambda \leq 0.5$. This result reveals that KNN compensates for the performance deficiencies of $f(\cdot)$ on data where it is more likely to make a mistake. Therefore, combining KNN using the error-aware integration strategy in \textsc{Araida} leads to an overall improvement in annotation quality, hence reducing human correction effort.

\section{Conclusion}
\label{conclusionref}
In interactive data annotation, an annotation model suggests labels to human annotators to verify. However, the annotation model is prone to errors when trained on limited labeled data. To tackle this challenge, we proposed \textsc{Araida}, an approach inspired by analogical reasoning, to compensate for the performance deficiencies of the annotation model and correct its mistakes using an error-aware integration strategy. Extensive experiments demonstrated that \textsc{Araida} is flexible to combine with different annotation models across various tasks and yields consistent improvement in label suggestion accuracy, which leads to a reduction of human correction effort. 
In this study, our method explores a new solution to bring more flexibility by allowing the human to design any preferred annotation model according to different annotation tasks. We are devoted to optimizing human-machine utilities by emphasizing the learning of task-specified concepts efficiently from a few human demonstrations.
In future work, we plan to extend \textsc{Araida} to other annotation tasks and develop it as a general toolkit that can benefit the NLP community.

\section{Acknowledgements}
This work was supported in part by the National Natural Science Foundation of China (No. 62272330); in part by the Fundamental Research Funds
for the Central Universities (No. YJ202219).

\FloatBarrier

\section{Limitations}

\paragraph{Human Studies.} This work aims to reduce human correction effort in interactive data annotation. We follow previous work~\citep{hwa2000sample,kristjansson2004interactive} to use the number of model suggestion errors to approximate the human correction effort needed. However, the actual effort needed depends on the particular example and the type of errors (e.g., whether it is obvious). Ideally, we would involve human annotators and measure the saving of annotation time. However, due to the large number of experimental settings, conducting human studies with each annotation model and ablation baseline was infeasible.

\paragraph{Error-Prone Human Annotation.} This paper treats human annotations as ground truth following previous studies in interactive data annotation~\citep{klie-etal-2018-inception,le2021interactive}. However, uncertainty and inconsistency of human annotations do occur. We refer readers to the literature on handling error-prone human annotation, such as crowd-sourced data annotation~\citep{larson2020inconsistencies}. 

Although human annotation errors are not the focus of this work, we explore \textsc{Araida}'s performance under synthesized label noise conditions. We consider the crowd-sourced data annotation scenario and assume that each human annotator $h_i$ makes mistakes with the latent probability $p_e^i \sim (0,0.3)$. We set the total number of annotators $O=10$ and sample their corresponding error probabilities $P_e=\{p_e^1, p_e^2, ..., p_e^O\}$. Then, we sample $u_i$ from a uniform distribution $U(0,1)$ for each annotation. If $u_i \leq p_e^i$, $h_i$ assigns a randomly sampled incorrect label; otherwise, it assigns the correct one. We use majority voting of the 10 annotators to obtain the final annotations following \citet{shirani2019learning}.

\begin{figure}[htb]
\centering
\includegraphics[width=0.25\textwidth]{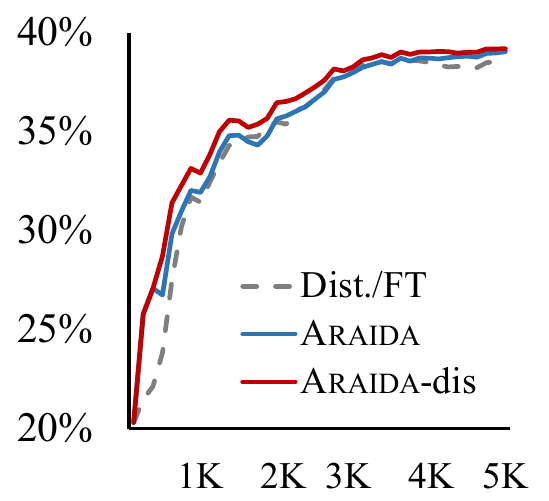}
\caption{MAC scores of various methods with synthesized label noise on the SST-5 dataset. Dist./FT is used as the annotation model. \textsc{Araida}-dis refers to \textsc{Araida} with a modified datastore maintenance strategy.}
\vspace{-4mm}
\label{fig:proofdwe2ddddd3}
\end{figure}

We slightly modified \textsc{Araida}'s datastore maintenance strategy. When the datastore exceeds its budge, we discard the data from the majority class and most \textbf{dis-similar} to its class prototype instead of removing the most similar one (as in the original \textsc{Araida} strategy). This strategy may help remove incorrectly labeled data. The experimental result in Figure~\ref{fig:proofdwe2ddddd3} shows that \textsc{Araida} still outperforms the baseline. The modified datastore maintenance strategy (\textsc{Araida}-dis) further improves the performance by a slight margin. Further research and more rigorous experiments are required to address the human annotation noise problem in interactive annotation.

\paragraph{Latency.} The KNN component requires retrieving similar examples as the input data, which may limit our method's time efficiency when the datastore size is large. To address this problem, besides the proposed datastore management strategy, we can also employ an efficient similarity search library such as FAISS~\footnote{\url{https://github.com/facebookresearch/faiss}} to speed up the retrieval.

\bibliography{custom}

\begin{thebibliography}{55}
\expandafter\ifx\csname natexlab\endcsname\relax\def\natexlab#1{#1}\fi

\bibitem[{Asai et~al.(2023)Asai, Min, Zhong, and Chen}]{asai-etal-2023-retrieval}
Akari Asai, Sewon Min, Zexuan Zhong, and Danqi Chen. 2023.
\newblock \href {https://doi.org/10.18653/v1/2023.acl-tutorials.6} {Retrieval-based language models and applications}.
\newblock In \emph{Proceedings of the 61st Annual Meeting of the Association for Computational Linguistics (Volume 6: Tutorial Abstracts)}, pages 41--46, Toronto, Canada. Association for Computational Linguistics.

\bibitem[{Bautista et~al.(2016)Bautista, Sanakoyeu, Tikhoncheva, and Ommer}]{NIPS2016_65fc52ed}
Miguel~A Bautista, Artsiom Sanakoyeu, Ekaterina Tikhoncheva, and Bjorn Ommer. 2016.
\newblock \href {https://proceedings.neurips.cc/paper_files/paper/2016/file/65fc52ed8f88c81323a418ca94cec2ed-Paper.pdf} {Cliquecnn: Deep unsupervised exemplar learning}.
\newblock In \emph{Advances in Neural Information Processing Systems}, volume~29. Curran Associates, Inc.

\bibitem[{Bollacker et~al.(2008)Bollacker, Evans, Paritosh, Sturge, and Taylor}]{bollacker2008freebase}
Kurt Bollacker, Colin Evans, Praveen Paritosh, Tim Sturge, and Jamie Taylor. 2008.
\newblock Freebase: a collaboratively created graph database for structuring human knowledge.
\newblock In \emph{Proceedings of the 2008 ACM SIGMOD international conference on Management of data}, pages 1247--1250.

\bibitem[{Casanova et~al.(2020)Casanova, Pinheiro, Rostamzadeh, and Pal}]{casanova2020reinforced}
Arantxa Casanova, Pedro~O Pinheiro, Negar Rostamzadeh, and Christopher~J Pal. 2020.
\newblock Reinforced active learning for image segmentation.
\newblock \emph{arxiv}.

\bibitem[{Chaudhary et~al.(2021)Chaudhary, Anastasopoulos, Sheikh, and Neubig}]{chaudhary2021reducing}
Aditi Chaudhary, Antonios Anastasopoulos, Zaid Sheikh, and Graham Neubig. 2021.
\newblock Reducing confusion in active learning for part-of-speech tagging.
\newblock \emph{TACL}, 9:1--16.

\bibitem[{Chen et~al.(2020)Chen, Du, He, and Wang}]{chen2020jit2r}
Xu~Chen, Changying Du, Xiuqiang He, and Jun Wang. 2020.
\newblock Jit2r: A joint framework for item tagging and tag-based recommendation.
\newblock In \emph{Proceedings of the 43rd international ACM SIGIR conference on research and development in information retrieval}, pages 1681--1684.

\bibitem[{Dasgupta(2005)}]{dasgupta2005coarse}
Sanjoy Dasgupta. 2005.
\newblock Coarse sample complexity bounds for active learning.
\newblock In \emph{Proceedings of the 18th International Conference on Neural Information Processing Systems}, pages 235--242.

\bibitem[{Dekel et~al.(2005)Dekel, Shalev-Shwartz, and Singer}]{dekel2005forgetron}
Ofer Dekel, Shai Shalev-Shwartz, and Yoram Singer. 2005.
\newblock The forgetron: A kernel-based perceptron on a fixed budget.
\newblock \emph{NeurIPS}, 18.

\bibitem[{Desmond et~al.(2021)Desmond, Duesterwald, Brimijoin, Brachman, and Pan}]{pmlr-v133-desmond21a}
Michael Desmond, Evelyn Duesterwald, Kristina Brimijoin, Michelle Brachman, and Qian Pan. 2021.
\newblock \href {https://proceedings.mlr.press/v133/desmond21a.html} {Semi-automated data labeling}.
\newblock In \emph{Proceedings of the NeurIPS 2020 Competition and Demonstration Track}, volume 133 of \emph{Proceedings of Machine Learning Research}, pages 156--169. PMLR.

\bibitem[{Dettmers et~al.(2018)Dettmers, Minervini, Stenetorp, and Riedel}]{dettmers2018convolutional}
Tim Dettmers, Pasquale Minervini, Pontus Stenetorp, and Sebastian Riedel. 2018.
\newblock Convolutional 2d knowledge graph embeddings.
\newblock In \emph{Proceedings of the AAAI conference on artificial intelligence}, volume~32.

\bibitem[{Dou et~al.(2019)Dou, Yu, and Anastasopoulos}]{dou2019investigating}
Zi-Yi Dou, Keyi Yu, and Antonios Anastasopoulos. 2019.
\newblock Investigating meta-learning algorithms for low-resource natural language understanding tasks.
\newblock In \emph{EMNLP}, pages 1192--1197.

\bibitem[{Gilardi et~al.(2023)Gilardi, Alizadeh, and Kubli}]{gilardi2023chatgpt}
Fabrizio Gilardi, Meysam Alizadeh, and Ma{\"e}l Kubli. 2023.
\newblock Chatgpt outperforms crowd-workers for text-annotation tasks.
\newblock \emph{arXiv preprint arXiv:2303.15056}.

\bibitem[{He et~al.(2023)He, Lin, Gong, Zhang, Lin, Jiao, Yiu, Duan, Chen et~al.}]{he2023annollm}
Xingwei He, Zhenghao Lin, Yeyun Gong, Hang Zhang, Chen Lin, Jian Jiao, Siu~Ming Yiu, Nan Duan, Weizhu Chen, et~al. 2023.
\newblock Annollm: Making large language models to be better crowdsourced annotators.
\newblock \emph{arXiv preprint arXiv:2303.16854}.

\bibitem[{Hedderich et~al.(2021)Hedderich, Lange, and Klakow}]{hedderich2021anea}
Michael~A Hedderich, Lukas Lange, and Dietrich Klakow. 2021.
\newblock Anea: distant supervision for low-resource named entity recognition.
\newblock \emph{the Practical Machine Learning For Developing Countries Workshop at ICLR}.

\bibitem[{Huang et~al.(2024)Huang, Deng, Lei, Lv, and Dagan}]{huang2024selective}
Chen Huang, Yang Deng, Wenqiang Lei, Jiancheng Lv, and Ido Dagan. 2024.
\newblock \href {http://arxiv.org/abs/2405.12081} {Selective annotation via data allocation: These data should be triaged to experts for annotation rather than the model}.

\bibitem[{Huang et~al.(2023)Huang, Qin, Lei, and Lv}]{huang-etal-2023-reduce}
Chen Huang, Peixin Qin, Wenqiang Lei, and Jiancheng Lv. 2023.
\newblock \href {https://doi.org/10.18653/v1/2023.emnlp-main.670} {Reduce human labor on evaluating conversational information retrieval system: A human-machine collaboration approach}.
\newblock In \emph{Proceedings of the 2023 Conference on Empirical Methods in Natural Language Processing}, pages 10876--10891, Singapore. Association for Computational Linguistics.

\bibitem[{Hwa(2000)}]{hwa2000sample}
Rebecca Hwa. 2000.
\newblock Sample selection for statistical grammar induction.
\newblock In \emph{2000 Joint SIGDAT Conference on Empirical Methods in Natural Language Processing and Very Large Corpora}, pages 45--52.

\bibitem[{Jang et~al.(2016)Jang, Gu, and Poole}]{jang2016categorical}
Eric Jang, Shixiang Gu, and Ben Poole. 2016.
\newblock Categorical reparameterization with gumbel-softmax.
\newblock \emph{arxiv}.

\bibitem[{Jiang et~al.(2021)Jiang, Wang, Cao, Cheng, Huang, and Li}]{jiang-etal-2021-learning}
Qingnan Jiang, Mingxuan Wang, Jun Cao, Shanbo Cheng, Shujian Huang, and Lei Li. 2021.
\newblock \href {https://doi.org/10.18653/v1/2021.emnlp-main.579} {Learning kernel-smoothed machine translation with retrieved examples}.
\newblock In \emph{Proceedings of the 2021 Conference on Empirical Methods in Natural Language Processing}, pages 7280--7290, Online and Punta Cana, Dominican Republic. Association for Computational Linguistics.

\bibitem[{Joulin et~al.(2017)Joulin, Grave, Bojanowski, and Mikolov}]{joulin2017bag}
Armand Joulin, Edouard Grave, Piotr Bojanowski, and Tomas Mikolov. 2017.
\newblock Bag of tricks for efficient text classification.
\newblock In \emph{EACL}, pages 427--431. Association for Computational Linguistics.

\bibitem[{Kassner and Sch{\"u}tze(2020)}]{kassner-schutze-2020-bert}
Nora Kassner and Hinrich Sch{\"u}tze. 2020.
\newblock \href {https://doi.org/10.18653/v1/2020.findings-emnlp.307} {{BERT}-k{NN}: Adding a k{NN} search component to pretrained language models for better {QA}}.
\newblock In \emph{Findings of the Association for Computational Linguistics: EMNLP 2020}, pages 3424--3430, Online. Association for Computational Linguistics.

\bibitem[{Khandelwal et~al.(2021)Khandelwal, Fan, Jurafsky, Zettlemoyer, and Lewis}]{khandelwal2020nearest}
Urvashi Khandelwal, Angela Fan, Dan Jurafsky, Luke Zettlemoyer, and Mike Lewis. 2021.
\newblock Nearest neighbor machine translation.
\newblock \emph{International Conference on Learning Representations}.

\bibitem[{Khandelwal et~al.(2019)Khandelwal, Levy, Jurafsky, Zettlemoyer, and Lewis}]{khandelwal2019generalization}
Urvashi Khandelwal, Omer Levy, Dan Jurafsky, Luke Zettlemoyer, and Mike Lewis. 2019.
\newblock Generalization through memorization: Nearest neighbor language models.
\newblock In \emph{ICLR}.

\bibitem[{Klie et~al.(2018)Klie, Bugert, Boullosa, Eckart~de Castilho, and Gurevych}]{klie-etal-2018-inception}
Jan-Christoph Klie, Michael Bugert, Beto Boullosa, Richard Eckart~de Castilho, and Iryna Gurevych. 2018.
\newblock The {INCE}p{TION} platform: Machine-assisted and knowledge-oriented interactive annotation.
\newblock In \emph{COLING: System Demonstrations}, pages 5--9, Santa Fe, New Mexico. ACL.

\bibitem[{Klie et~al.(2020)Klie, Eckart~de Castilho, and Gurevych}]{klie-etal-2020-zero}
Jan-Christoph Klie, Richard Eckart~de Castilho, and Iryna Gurevych. 2020.
\newblock {F}rom {Z}ero to {H}ero: {H}uman-{I}n-{T}he-{L}oop {E}ntity {L}inking in {L}ow {R}esource {D}omains.
\newblock In \emph{ACL}, pages 6982--6993, Online. ACL.

\bibitem[{{Kober} et~al.(2021){Kober}, {Weeds}, {Bertolini}, and {Weir}}]{kober2021data}
Thomas {Kober}, Julie {Weeds}, Lorenzo {Bertolini}, and David~J. {Weir}. 2021.
\newblock Data augmentation for hypernymy detection.
\newblock In \emph{EACL}, pages 1034--1048.

\bibitem[{Kristjansson et~al.(2004)Kristjansson, Culotta, Viola, and McCallum}]{kristjansson2004interactive}
Trausti Kristjansson, Aron Culotta, Paul Viola, and Andrew McCallum. 2004.
\newblock Interactive information extraction with constrained conditional random fields.
\newblock In \emph{AAAI}, volume~4, pages 412--418.

\bibitem[{Lake et~al.(2015)Lake, Salakhutdinov, and Tenenbaum}]{lake2015human}
Brenden~M Lake, Ruslan Salakhutdinov, and Joshua~B Tenenbaum. 2015.
\newblock Human-level concept learning through probabilistic program induction.
\newblock \emph{Science}, 350(6266):1332--1338.

\bibitem[{Lake et~al.(2017)Lake, Ullman, Tenenbaum, and Gershman}]{lake2017building}
Brenden~M Lake, Tomer~D Ullman, Joshua~B Tenenbaum, and Samuel~J Gershman. 2017.
\newblock Building machines that learn and think like people.
\newblock \emph{Behavioral and brain sciences}, 40.

\bibitem[{Larson et~al.(2020)Larson, Cheung, Mahendran, Leach, and Kummerfeld}]{larson2020inconsistencies}
Stefan Larson, Adrian Cheung, Anish Mahendran, Kevin Leach, and Jonathan~K Kummerfeld. 2020.
\newblock Inconsistencies in crowdsourced slot-filling annotations: A typology and identification methods.
\newblock In \emph{Proceedings of the 28th International Conference on Computational Linguistics}, pages 5035--5046.

\bibitem[{Laws et~al.(2011)Laws, Scheible, and Sch{\"u}tze}]{laws-etal-2011-active}
Florian Laws, Christian Scheible, and Hinrich Sch{\"u}tze. 2011.
\newblock Active learning with {A}mazon {M}echanical {T}urk.
\newblock In \emph{EMNLP}, pages 1546--1556, Edinburgh, Scotland, UK. ACL.

\bibitem[{Le et~al.(2021)Le, Nguyen, Tran, Nguyen, Hoang, Le, and Tran}]{le2021interactive}
Trung-Nghia Le, Tam~V Nguyen, Quoc-Cuong Tran, Lam Nguyen, Trung-Hieu Hoang, Minh-Quan Le, and Minh-Triet Tran. 2021.
\newblock Interactive video object mask annotation.
\newblock In \emph{AAAI}, volume~35, pages 16067--16070.

\bibitem[{Lee et~al.(2022)Lee, Chung, Kim, Song, and Kim}]{lee2022promptiverse}
Yoonjoo Lee, John Joon~Young Chung, Tae~Soo Kim, Jean~Y Song, and Juho Kim. 2022.
\newblock Promptiverse: Scalable generation of scaffolding prompts through human-ai hybrid knowledge graph annotation.
\newblock In \emph{CHI Conference on Human Factors in Computing Systems}, pages 1--18.

\bibitem[{Li et~al.(2021)Li, Yu, Quangang, and Liu}]{li2021fitannotator}
Yanzeng Li, Bowen Yu, Li~Quangang, and Tingwen Liu. 2021.
\newblock Fitannotator: A flexible and intelligent text annotation system.
\newblock In \emph{Proceedings of the 2021 Conference of the North American Chapter of the Association for Computational Linguistics: Human Language Technologies: Demonstrations}, pages 35--41.

\bibitem[{Liu et~al.(2023)Liu, Liu, Wong, Li, Jiao, Chao, and Zhang}]{liu2023knn}
Shudong Liu, Xuebo Liu, Derek~F Wong, Zhaocong Li, Wenxiang Jiao, Lidia~S Chao, and Min Zhang. 2023.
\newblock knn-tl: k-nearest-neighbor transfer learning for low-resource neural machine translation.
\newblock In \emph{Proceedings of the 61st Annual Meeting of the Association for Computational Linguistics (Volume 1: Long Papers)}, pages 1878--1891.

\bibitem[{Maas et~al.(2011)Maas, Daly, Pham, Huang, Ng, and Potts}]{maas-EtAl:2011:ACL-HLT2011}
Andrew~L. Maas, Raymond~E. Daly, Peter~T. Pham, Dan Huang, Andrew~Y. Ng, and Christopher Potts. 2011.
\newblock \href {http://www.aclweb.org/anthology/P11-1015} {Learning word vectors for sentiment analysis}.
\newblock In \emph{Proceedings of the 49th Annual Meeting of the Association for Computational Linguistics: Human Language Technologies}, pages 142--150, Portland, Oregon, USA. Association for Computational Linguistics.

\bibitem[{Marcos-Pablos and Garc{\'\i}a-Pe{\~n}alvo(2020)}]{marcos2020information}
Samuel Marcos-Pablos and Francisco~J Garc{\'\i}a-Pe{\~n}alvo. 2020.
\newblock Information retrieval methodology for aiding scientific database search.
\newblock \emph{Soft Computing}, 24(8):5551--5560.

\bibitem[{Mitchell(2021)}]{mitchell2021abstraction}
Melanie Mitchell. 2021.
\newblock Abstraction and analogy-making in artificial intelligence.
\newblock \emph{Annals of the New York Academy of Sciences}, 1505(1):79--101.

\bibitem[{Neelakantan et~al.(2022)Neelakantan, Xu, Puri, Radford, Han, Tworek, Yuan, Tezak, Kim, Hallacy et~al.}]{neelakantan2022text}
Arvind Neelakantan, Tao Xu, Raul Puri, Alec Radford, Jesse~Michael Han, Jerry Tworek, Qiming Yuan, Nikolas Tezak, Jong~Wook Kim, Chris Hallacy, et~al. 2022.
\newblock Text and code embeddings by contrastive pre-training.
\newblock \emph{arXiv preprint arXiv:2201.10005}.

\bibitem[{Ouyang et~al.(2022)Ouyang, Wu, Jiang, Almeida, Wainwright, Mishkin, Zhang, Agarwal, Slama, Ray et~al.}]{ouyang2022training}
Long Ouyang, Jeffrey Wu, Xu~Jiang, Diogo Almeida, Carroll Wainwright, Pamela Mishkin, Chong Zhang, Sandhini Agarwal, Katarina Slama, Alex Ray, et~al. 2022.
\newblock Training language models to follow instructions with human feedback.
\newblock \emph{Advances in Neural Information Processing Systems}, 35:27730--27744.

\bibitem[{Pennington et~al.(2014)Pennington, Socher, and Manning}]{pennington2014glove}
Jeffrey Pennington, Richard Socher, and Christopher~D. Manning. 2014.
\newblock Glove: Global vectors for word representation.
\newblock In \emph{EMNLP}, pages 1532--1543.

\bibitem[{Rietz and Maedche(2021)}]{rietz2021cody}
Tim Rietz and Alexander Maedche. 2021.
\newblock Cody: An ai-based system to semi-automate coding for qualitative research.
\newblock In \emph{CHI}, pages 1--14.

\bibitem[{Ringger et~al.(2007)Ringger, McClanahan, Haertel, Busby, Carmen, Carroll, Seppi, and Lonsdale}]{ringger2007active}
Eric Ringger, Peter McClanahan, Robbie Haertel, George Busby, Marc Carmen, James Carroll, Kevin Seppi, and Deryle Lonsdale. 2007.
\newblock Active learning for part-of-speech tagging: Accelerating corpus annotation.
\newblock In \emph{Proceedings of the Linguistic Annotation Workshop}, pages 101--108.

\bibitem[{{Roller} et~al.(2014){Roller}, {Erk}, and {Boleda}}]{roller2014inclusive}
Stephen {Roller}, Katrin {Erk}, and Gemma {Boleda}. 2014.
\newblock Inclusive yet selective: Supervised distributional hypernymy detection.
\newblock In \emph{COLING: Technical Papers}, pages 1025--1036.

\bibitem[{Shirani et~al.(2019)Shirani, Dernoncourt, Asente, Lipka, Kim, Echevarria, and Solorio}]{shirani2019learning}
Amirreza Shirani, Franck Dernoncourt, Paul Asente, Nedim Lipka, Seokhwan Kim, Jose Echevarria, and Thamar Solorio. 2019.
\newblock Learning emphasis selection for written text in visual media from crowd-sourced label distributions.
\newblock In \emph{Proceedings of the 57th Annual Meeting of the Association for Computational Linguistics}, pages 1167--1172.

\bibitem[{Socher et~al.(2013)Socher, Perelygin, Wu, Chuang, Manning, Ng, and Potts}]{socher2013recursive}
Richard Socher, Alex Perelygin, Jean Wu, Jason Chuang, Christopher~D Manning, Andrew~Y Ng, and Christopher Potts. 2013.
\newblock Recursive deep models for semantic compositionality over a sentiment treebank.
\newblock In \emph{Proceedings of the 2013 conference on empirical methods in natural language processing}, pages 1631--1642.

\bibitem[{Touvron et~al.(2023)Touvron, Martin, Stone, Albert, Almahairi, Babaei, Bashlykov, Batra, Bhargava, Bhosale et~al.}]{touvron2023llama}
Hugo Touvron, Louis Martin, Kevin Stone, Peter Albert, Amjad Almahairi, Yasmine Babaei, Nikolay Bashlykov, Soumya Batra, Prajjwal Bhargava, Shruti Bhosale, et~al. 2023.
\newblock Llama 2: Open foundation and fine-tuned chat models.
\newblock \emph{arXiv preprint arXiv:2307.09288}.

\bibitem[{Wang et~al.(2022{\natexlab{a}})Wang, Prabhat, and Sambasivan}]{wang2022whose}
Ding Wang, Shantanu Prabhat, and Nithya Sambasivan. 2022{\natexlab{a}}.
\newblock Whose ai dream? in search of the aspiration in data annotation.
\newblock In \emph{CHI}, pages 1--16.

\bibitem[{Wang et~al.(2021)Wang, Wei, Zhang, Huang, Xie, Luo, and Chen}]{wang2021non}
Dongqi Wang, Haoran Wei, Zhirui Zhang, Shujian Huang, Jun Xie, Weihua Luo, and Jiajun Chen. 2021.
\newblock Non-parametric online learning from human feedback for neural machine translation.
\newblock \emph{arXiv}.

\bibitem[{Wang et~al.(2022{\natexlab{b}})Wang, Li, Meng, Zhang, Ouyang, Li, and Wang}]{wang2022k}
Shuhe Wang, Xiaoya Li, Yuxian Meng, Tianwei Zhang, Rongbin Ouyang, Jiwei Li, and Guoyin Wang. 2022{\natexlab{b}}.
\newblock $ k $ nn-ner: Named entity recognition with nearest neighbor search.
\newblock \emph{arXiv preprint arXiv:2203.17103}.

\bibitem[{Wang et~al.(2019)Wang, Chao, Weinberger, and van~der Maaten}]{wang2019simpleshot}
Yan Wang, Wei-Lun Chao, Kilian~Q Weinberger, and Laurens van~der Maaten. 2019.
\newblock Simpleshot: Revisiting nearest-neighbor classification for few-shot learning.
\newblock \emph{arXiv}.

\bibitem[{Wu et~al.(2022)Wu, Xiao, Sun, Zhang, Ma, and He}]{WU2022364}
Xingjiao Wu, Luwei Xiao, Yixuan Sun, Junhang Zhang, Tianlong Ma, and Liang He. 2022.
\newblock \href {https://doi.org/https://doi.org/10.1016/j.future.2022.05.014} {A survey of human-in-the-loop for machine learning}.
\newblock \emph{Future Generation Computer Systems}, 135:364--381.

\bibitem[{Xiao et~al.(2023)Xiao, Dong, Zhao, Wu, Lin, Chen, and Wang}]{xiao-etal-2023-freeal}
Ruixuan Xiao, Yiwen Dong, Junbo Zhao, Runze Wu, Minmin Lin, Gang Chen, and Haobo Wang. 2023.
\newblock \href {https://doi.org/10.18653/v1/2023.emnlp-main.896} {{F}ree{AL}: Towards human-free active learning in the era of large language models}.
\newblock In \emph{Proceedings of the 2023 Conference on Empirical Methods in Natural Language Processing}, pages 14520--14535, Singapore. Association for Computational Linguistics.

\bibitem[{Yao et~al.(2021)Yao, Qin, Zhu, Ma, Zhang, Du, and Xiong}]{yao2021interactive}
Kaichun Yao, Chuan Qin, Hengshu Zhu, Chao Ma, Jingshuai Zhang, Yi~Du, and Hui Xiong. 2021.
\newblock An interactive neural network approach to keyphrase extraction in talent recruitment.
\newblock In \emph{Proceedings of the 30th ACM International Conference on Information \& Knowledge Management}, pages 2383--2393.

\bibitem[{Zheng et~al.(2021)Zheng, Zhang, Guo, Huang, Chen, Luo, and Chen}]{zheng2021adaptive}
Xin Zheng, Zhirui Zhang, Junliang Guo, Shujian Huang, Boxing Chen, Weihua Luo, and Jiajun Chen. 2021.
\newblock Adaptive nearest neighbor machine translation.
\newblock pages 368--374.

\end{thebibliography}
\FloatBarrier

\appendix

\begin{figure*}[!htb]
\centering
\begin{tabular}{cccc}
\includegraphics[width=0.22\textwidth]{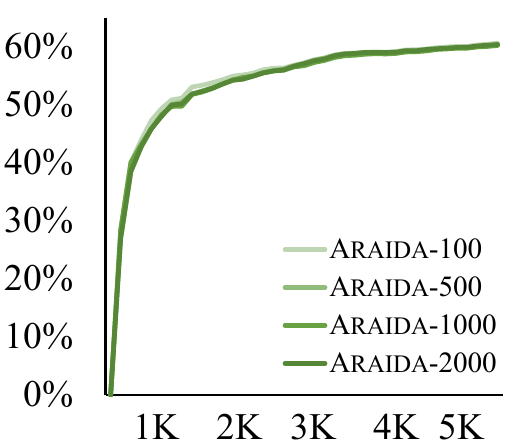}&
\includegraphics[width=0.22\textwidth]{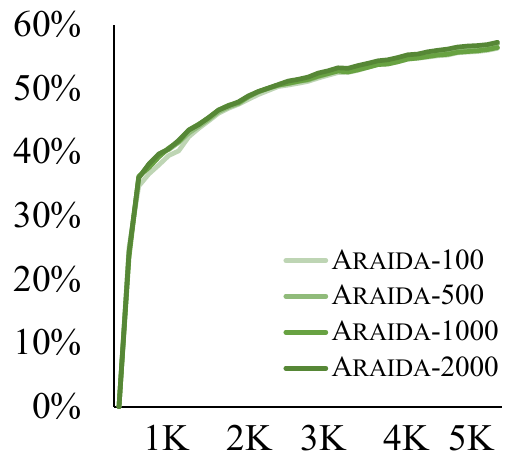}&
\includegraphics[width=0.22\textwidth]{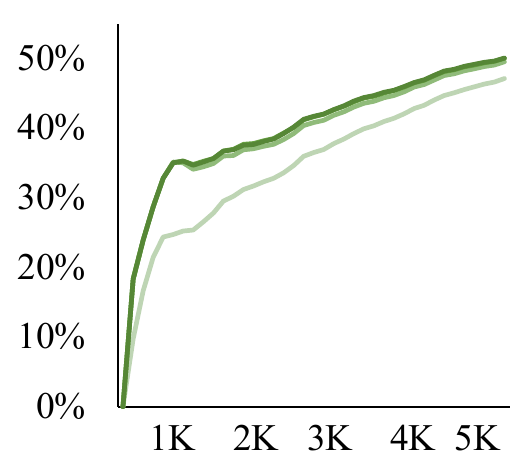}&
\includegraphics[width=0.22\textwidth]{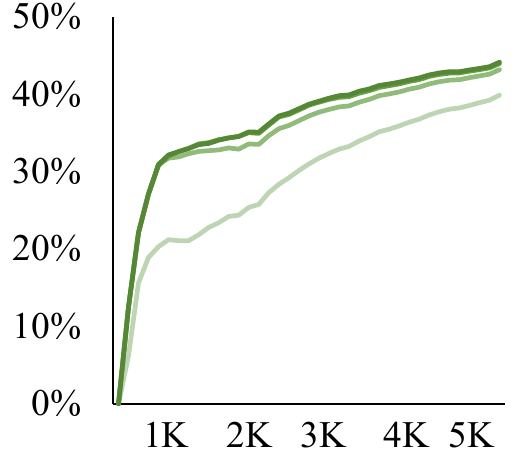}\\

(a) WN18RR &(b) WN18RR + AL & (c) FreeBase &(d) FreeBase + AL \\
\includegraphics[width=0.22\textwidth]{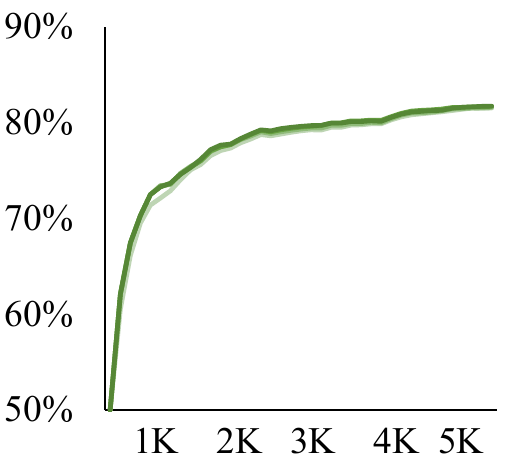}&
\includegraphics[width=0.22\textwidth]{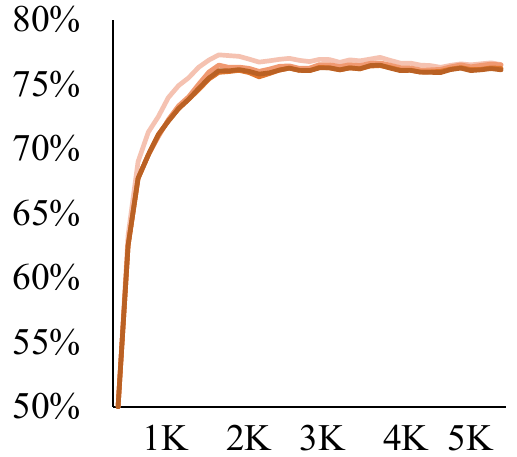} &
\includegraphics[width=0.22\textwidth]{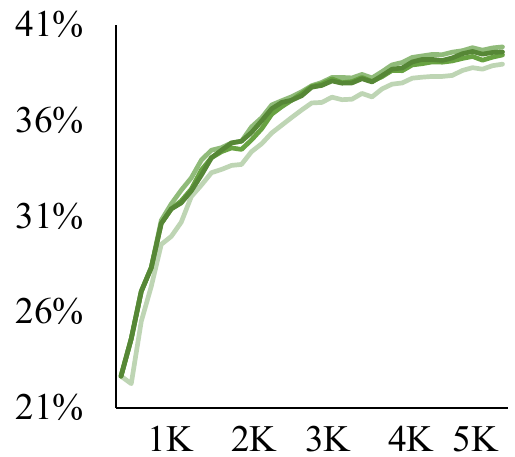}&
\includegraphics[width=0.22\textwidth]{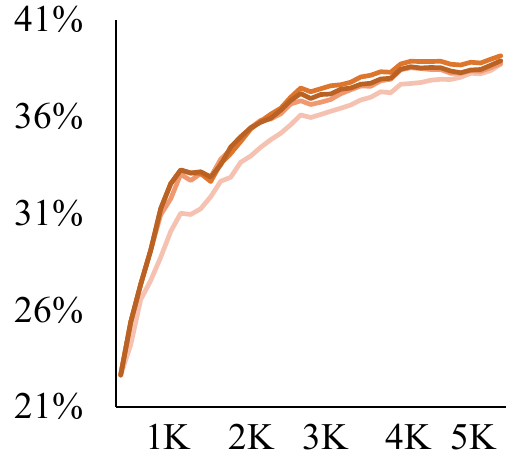}\\
(e) IMDB &(f) IMDB + AL & (g) SST-5 &(h) SST-5 + AL
\end{tabular}
\caption{Machine Cumulative Accuracy of \textsc{Araida} with different datastore sizes.}
\label{fig:proof2ddddd}
\end{figure*}

\begin{table*}[]
\centering
\resizebox{0.98\textwidth}{!}{%
\begin{tabular}{ll|ll}
\toprule
\multicolumn{2}{c|}{\textbf{WN18RR}} & \multicolumn{2}{c}{\textbf{FreeBase}} \\ \hline
\multicolumn{1}{l|}{\textbf{Raw Class}} & \textbf{Mapped Class} & \multicolumn{1}{l|}{\textbf{Raw Class}} & \textbf{Mapped Class} \\ \midrule
\multicolumn{1}{l|}{\_hypernym} & hypernym & \multicolumn{1}{l|}{/location/location/contains} & contains \\ \hline
\multicolumn{1}{l|}{\_derivationally\_related\_form} & derivation & \multicolumn{1}{l|}{/olympics/olympic\_sport/athletes./olympics/olympic\_athlete\_affiliation/country} & country \\ \hline
\multicolumn{1}{l|}{\_member\_meronym} & member & \multicolumn{1}{l|}{/music/performance\_role/track\_performances./music/track\_contribution/role} & track\_role \\ \hline
\multicolumn{1}{l|}{\_has\_part} & component & \multicolumn{1}{l|}{/people/person/profession} & profession \\ \hline
\multicolumn{1}{l|}{\_synset\_domain\_topic\_of} & synset & \multicolumn{1}{l|}{/music/performance\_role/regular\_performances./music/group\_membership/role} & group\_role \\ \hline
\multicolumn{1}{l|}{\_instance\_hypernym} & instance of hypernym & \multicolumn{1}{l|}{/location/location/adjoin\_s./location/adjoining\_relationship/adjoins} & adjoins \\ \hline
\multicolumn{1}{l|}{\_also\_see} & synonym & \multicolumn{1}{l|}{/film/film/release\_date\_s./film/film\_regional\_release\_date/film\_release\_region} & film\_release \\ \hline
\multicolumn{1}{l|}{\_verb\_group} & verb group & \multicolumn{1}{l|}{/food/food/nutrients./food/nutrition\_fact/nutrient} & nutrient \\ \bottomrule
\end{tabular}%
}
\caption{Class mapping details for WN18RR and FreeBase}
\label{tab:yes}
\end{table*}

\section{Hyperparameter Analysis}
\label{sec4}
In this section, we provide a detailed hyperparameter analysis of \textsc{Araida}, including the datastore size $A_t$, the number of neighbors $k$ for $\rho_t$, and the datastore maintenance strategy. We show results only for the Dist./FT annotation model for brevity. 

\subsection{Datastore Size}
\label{dyenicx}

We tune the size of the datastore $A_t$ in $\{100,500,1000,2000\}$ and evaluate the machine cumulative accuracy of \textsc{Araida} with or without active learning. We illustrate the results in Fig.\ref{fig:proof2ddddd}. A larger datastore size generally brings higher annotation performance since it allows us to maintain more data from past human-machine interactions. However, it also requires larger memory usage and causes longer latency. We found that a datastore size of 1000 is a reasonable tradeoff, which we utilize in our main experiments.

\subsection{Top $k$ for $\rho_t$}
\label{dyuebkru3e}
\begin{figure*}[!htb]
\centering
\begin{tabular}{cccc}
\includegraphics[width=0.22\textwidth]{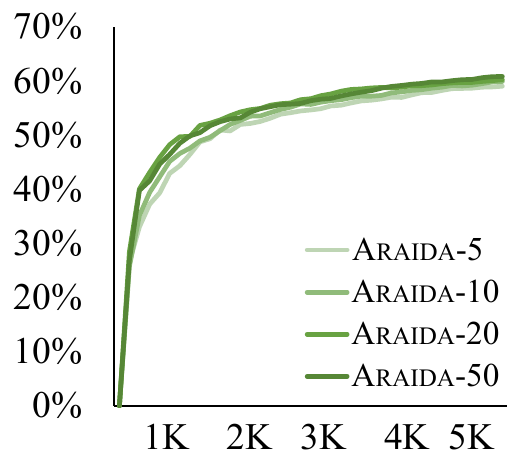}&
\includegraphics[width=0.22\textwidth]{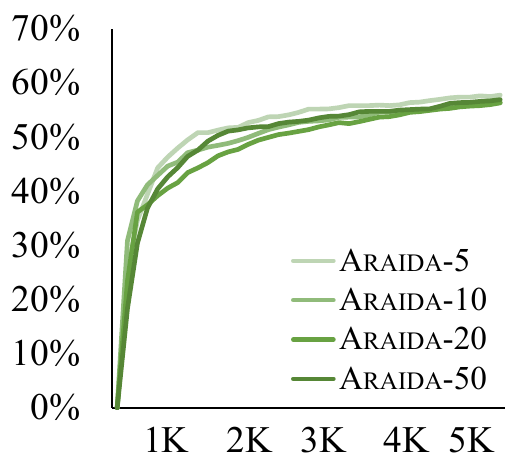}&
\includegraphics[width=0.22\textwidth]{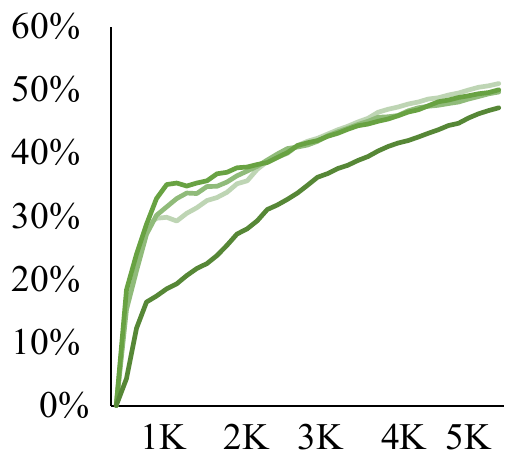}&
\includegraphics[width=0.22\textwidth]{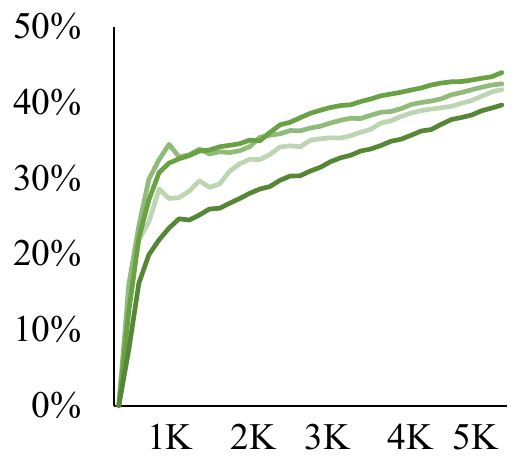}\\
(a) WN18RR &(b) WN18RR + AL&(c) FreeBase &(d) FreeBase + AL \\
\includegraphics[width=0.22\textwidth]{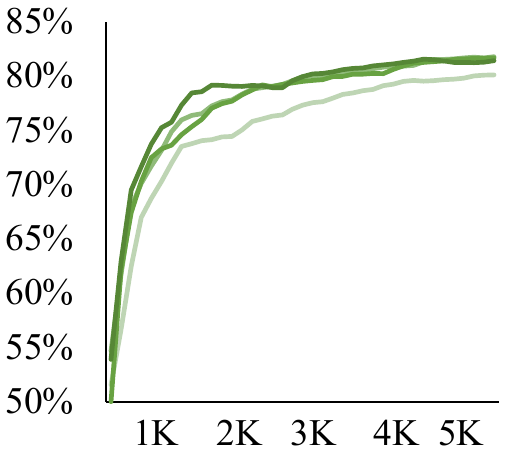}&
\includegraphics[width=0.22\textwidth]{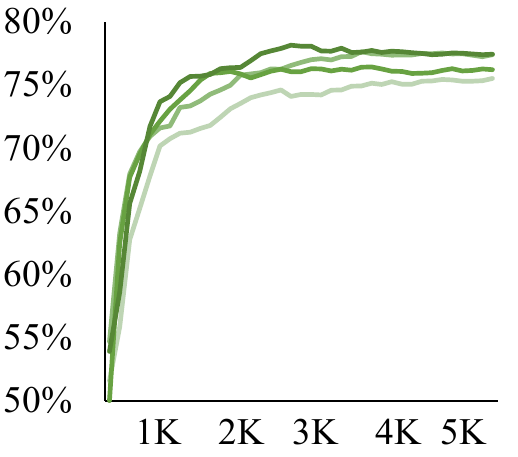} &
\includegraphics[width=0.22\textwidth]{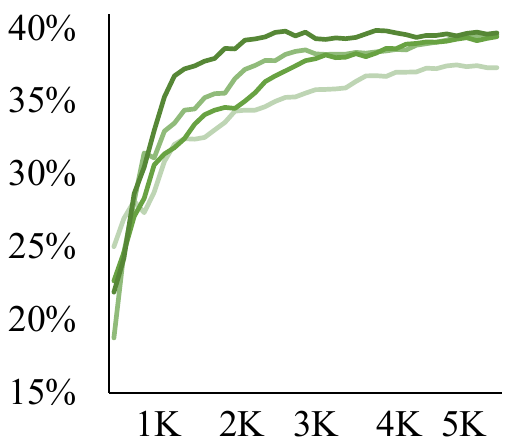}&
\includegraphics[width=0.22\textwidth]{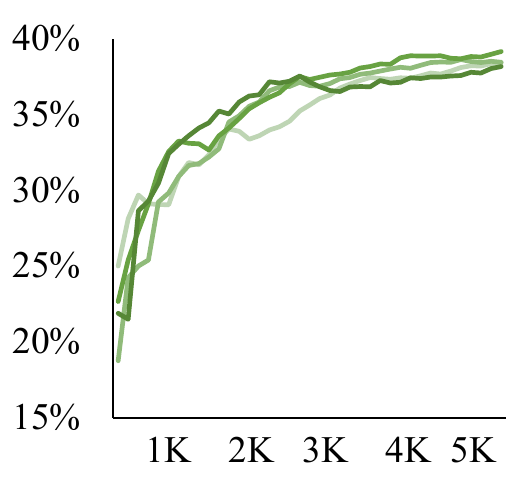}\\
(e) IMDB &(f) IMDB + AL & (g) SST-5 &(h) SST-5 + AL

\end{tabular}
\caption{Machine Cumulative Accuracy of \textsc{Araida} with different $k$ for $\rho_t$.}
\label{fig:proof2ddddd2}
\end{figure*}

We tune the number of neighbors $A_t$ in $\{5,10,20,50\}$ and evaluate the machine cumulative accuracy of \textsc{Araida} with or without active learning. As shown in Fig.\ref{fig:proof2ddddd2}, $k=20$ seems to perform well for all tasks.

\begin{table}[!htp]
\centering
\caption{Results on comparing to fully fine-tuned models}
\label{tab:my-tablehddh}
\resizebox{0.48\textwidth}{!}{%
\begin{tabular}{l|c|c|c|c}
\toprule
{ \textbf{Model}}            & { \textbf{WN18RR}} & { \textbf{FreeBase}} & { \textbf{IMDB}} & { \textbf{SST-5}} \\
\midrule
{ BERT(fully finetuned)}     & { 75.13$\pm$1.12}      & { 60.47$\pm$1.24}        & { 90.04$\pm$0.42}    & { 46.59$\pm$1.20}     \\
{ BERT+\textsc{Araida}}               & { 54.41$\pm$1.51}      & { 50.30$\pm$2.25}        & { 88.79$\pm$3.54}    & { 43.81$\pm$3.12}     \\ \hline
{ Dist./FT(fully finetuned)} & { 72.64$\pm$1.05}      & { 58.50$\pm$1.36}        & { 84.87$\pm$0.81}    & { 42.12$\pm$1.44}     \\
{ Dist./FT+\textsc{Araida}}           & { 52.16$\pm$1.37}      & { 43.02$\pm$6.43}        & { 79.33$\pm$2.81}    & { 37.21$\pm$3.03}    \\ \bottomrule
\end{tabular}%
}
\end{table}

\subsection{Datastore Maintenance Strategy}
\label{bufferthis}

\begin{figure*}[!htb]
\centering
\begin{tabular}{cccc}
\includegraphics[width=0.22\textwidth]{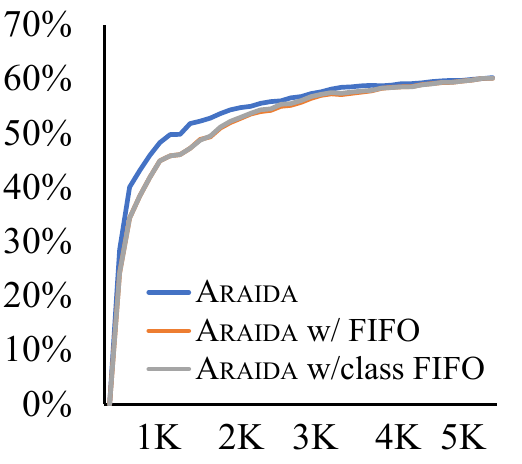}&
\includegraphics[width=0.22\textwidth]{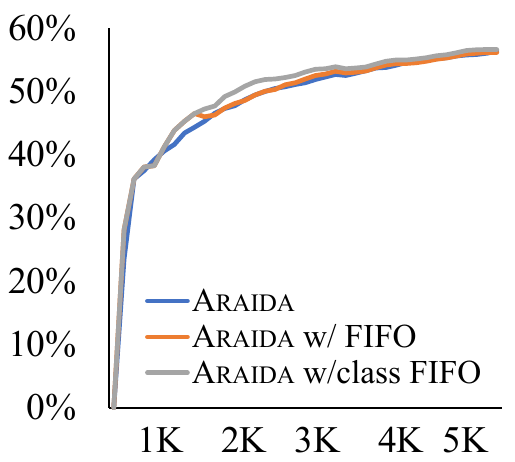}  & 
\includegraphics[width=0.22\textwidth]{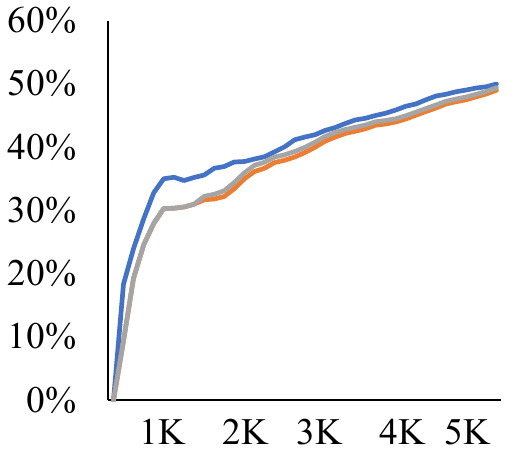}&
\includegraphics[width=0.22\textwidth]{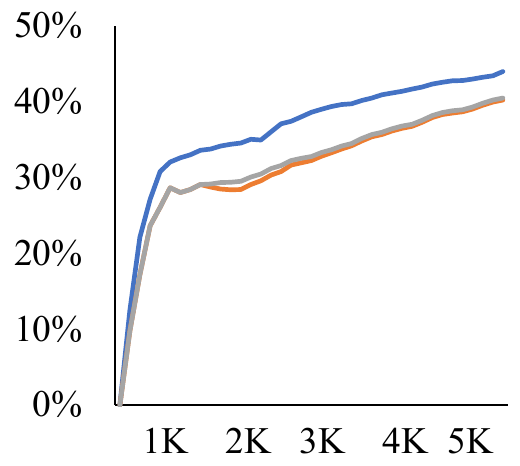}\\
(a) WN18RR &(b) WN18RR + AL & (c) FreeBase &(d) FreeBase + AL \\
\includegraphics[width=0.22\textwidth]{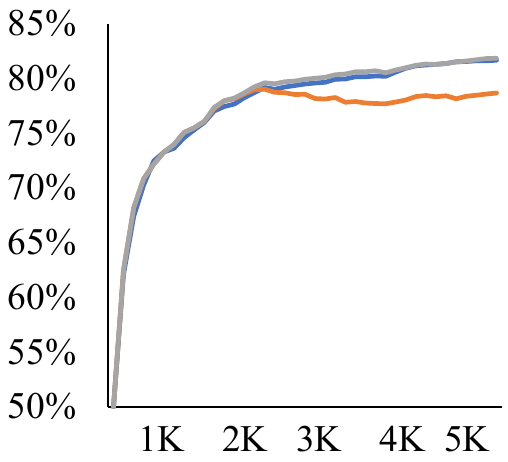}&
\includegraphics[width=0.22\textwidth]{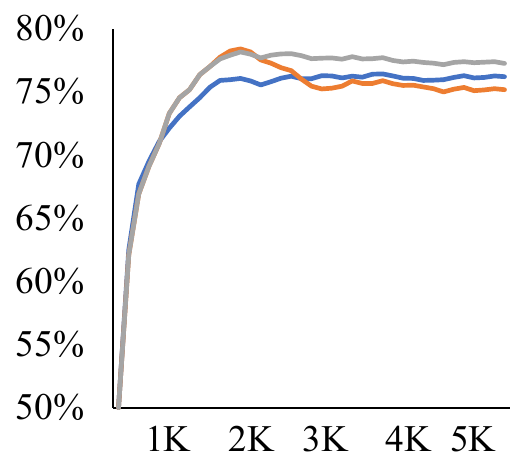} &
\includegraphics[width=0.22\textwidth]{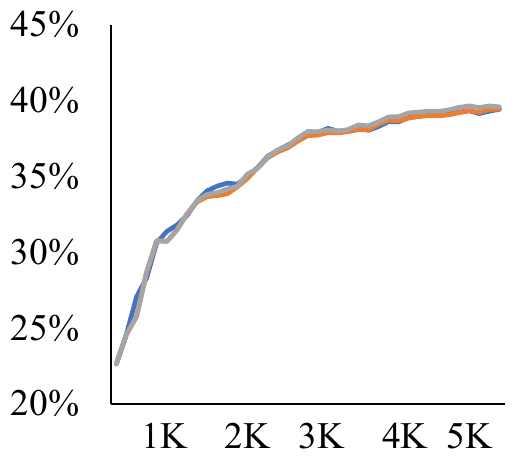}&
\includegraphics[width=0.22\textwidth]{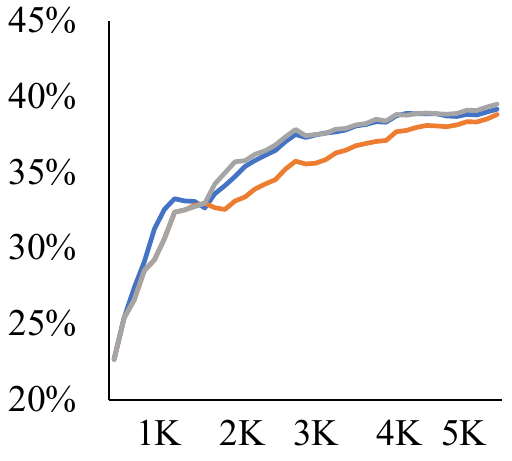}\\
(e) IMDB &(f) IMDB + AL &(g) SST-5 &(h) SST-5 + AL
\end{tabular}
\caption{Machine Cumulative Accuracy of \textsc{Araida} with different datastore maintenance strategies.}
\label{fig:proof2ddddd3}
\end{figure*}

We propose a class-aware datastore maintenance strategy for \textsc{Araida}, which removes labeled examples from the majority class most similar to their class prototype. Compared to the conventional First-In-First-Out (FIFO) strategy~\citep{dekel2005forgetron}, our method ensures that 1) the datastore contains as much data from different classes as possible, and 2) the class prototype is least affected after the removal. We compare with two variants of \textsc{Araida}, including \textit{\textsc{Araida} w/ FIFO} and \textit{\textsc{Araida} w/ class FIFO}. The former discards the oldest example regardless of the class; the latter discards the oldest example belonging to the majority class in the datastore.  

As shown in Fig.\ref{fig:proof2ddddd3}, \textit{\textsc{Araida} w/ FIFO} can suffer from a sudden decrease in performance, as it may remove important examples arbitrarily. After integrating the class information, \textit{\textsc{Araida} w/ class FIFO} removes the oldest analogy from the majority class, achieving a comparative performance to \textsc{Araida}. However, \textsc{Araida} still performs better when the number of classes increases (e.g., FreeBase).

\section{Results on Comparing to Fully Fine-tuned Model}
We utilized up to 5K of data to simulate the limited data annotation task, and we updated the parameters of the annotation model as the interactive annotation process progressed. To validate the effectiveness of the fully fine-tuned model, we use the remaining data from the original dataset, excluding the 5K data and the validation data, as the training data to fully fine-tune the model. Next, we employ this model for interactive annotation. Since it is already fully fine-tuned, we do not update the model parameters during the annotation process. Due to time constraints, we are currently only considering scenarios where active learning has not been adopted. Additionally, we only fine-tune small annotation models (i.e., BERT and Dist./FT). 

Based on the results in the Table \ref{tab:my-tablehddh}, it is evident that the annotation performance of the fully fine-tuned model surpasses that of our method. This suggests that while our method currently offers a lightweight approach to aid data annotation, which can to some extent enhance the sample efficiency of the annotation model, there is still potential for improvement in future research.Nevertheless, it's important to emphasize that in real-world data annotation scenarios, we typically don't have access to fully fine-tuned models initially, as we lack labeled datasets. While one might turn to LLMs like ChatGPT, our findings in Figure \ref{main_results2} indicate that our method could potentially improve upon ChatGPT even further.

\section{Results on Smaller-sized Transformer-based Model}
With a wide array of language models to choose from, we faced the challenge of not being able to test all available models. To address this, we selected three prominent models (GloVe, LLaMa2, and ChatGPT) based on our hardware resource capabilities in our main experiments. While open to conducting further experiments, our focus was limited by time constraints, leading us to concentrate solely on fine-tuning BERT (w/o AL), with the outcomes detailed in Table \ref{tab:my-tablebert}. After implementing \textsc{Araida}, we observed a significant enhancement in the quality of annotations by using our \textsc{Araida}.

\begin{table}[!ht]
\centering
\caption{Results on smaller-sized Transformer-based model}
\label{tab:my-tablebert}
\resizebox{0.49\textwidth}{!}{%
\begin{tabular}{l|c|c|c|c}
\toprule
\textbf{Model}      & \textbf{WN18RR}     & \textbf{FreeBase}   & \textbf{IMDB}       & \textbf{SST-5}      \\ \midrule
BERT                & 53.32$\pm$1.02          & 46.12$\pm$3.23          & 85.38$\pm$7.16          & 41.08$\pm$3.41          \\ \hline
BERT+\textsc{Araida}         & \textbf{54.41$\pm$1.51} & \textbf{50.30$\pm$2.25} & \textbf{88.79$\pm$3.54} & \textbf{43.81$\pm$3.12} \\
BERT+\textsc{Araida} w/o f & 51.03$\pm$2.11          & 44.63$\pm$4.04          & 80.29$\pm$1.09          & 38.51$\pm$1.45         \\ \bottomrule
\end{tabular}%
}
\end{table}

\section{Prompts and Fine-Tuning Examples for LLM-Based Annotation Models} 
\label{apendix:llm}

Table~\ref{tab:prompt} presents prompts used for zero-shot and few-shot annotation. Table~\ref{tab:llama} shows fine-tuning examples for LLaMa2$_{sft}$.

\begin{table*}[]
\centering
\resizebox{0.98\textwidth}{!}{%
\begin{tabular}{l|l}
\toprule
\textbf{Dataset} & \textbf{Prompts} \\ \midrule
IMDB & \begin{tabular}[c]{@{}l@{}}==================\\ input: If you like original gut wrenching laughter...it. Great Camp!!!.\\ output: positive\\ input: I saw this movie when I was about 12 when it ... There are no rules.\\ output: negative \\ ==================\\ You need to identify the sentiments of the following sentences, output positive or negative.\\ \{INPUTS\}\end{tabular} \\ \hline
SST-5 & \begin{tabular}[c]{@{}l@{}}=====================\\ input: The gorgeously elaborate continuation ...Tolkien 's Middle-earth.\\ output: Strong Positive\\ input: Singer/composer Bryan Adams contributes a slew of..., spirit of the piece.\\ output: Positive\\ input: You'd think by now ...with hearts of gold.\\ output: Neutral\\ input: This isn't a new idea .\\ output: Negative\\ input: A sour little movie at ... What was it all for ? \\ output: Strong Negative\\ ====================\\ Identify the sentiment of each paragraph, \\ you have five options: 'Strong Positive', 'Positive', 'Neutral', 'Negative' or 'Strong Negative'\\ \{INPUTS\}\end{tabular} \\ \hline
WN18RR & \begin{tabular}[c]{@{}l@{}}=============\\ input: ability unfitness\\ output: antonym\\ input: dissent debating\\ output: entailment\\ ...\\ input: abandonment apostasy\\ output: hypernym\\ input: abandonment discard\\ output: hyponym\\ input: Afghanistan Afghan\\ output: member\\ input: abandonment abandonment\\ output: synonym\\ =============\\ Identify the semantic relation of the each word pair, \\ you have eight options: 'component', 'synset',..., 'hypernym', 'derivation', 'member', 'synonym'\\ You MUST only output the semantic relation word for each input!\\ \{INPUTS\}\end{tabular} \\ \hline
FreeBase & \begin{tabular}[c]{@{}l@{}}========== \\ input: Libya Egypt \\ output: adjoins \\ input: Honolulu Punahou \\ output: contains \\ input: Bobsleigh Netherlands \\ output: country \\ input: Blackbriar Lithuania \\ output: film\_release \\ input: Autoharp Guitar \\ output: group\_role \\ input: IceCream Water \\ output: nutrient \\ input: Shriya Actor \\ output: profession \\ input: Cello Pennywhistle \\ output: track\_role \\ ========= \\ Identify the semantic relation of the each word pair, \\ you have eight options: 'contains', 'country', 'track\_role', 'profession', 'group\_role', 'adjoins', 'film\_release', 'nutrient'.\\ only output the semantic relation.\\ \{INPUTS\}\end{tabular} \\ \bottomrule
\end{tabular}%
}
\caption{Prompts for different datasets to obtain the annotation. We remove the few-shot demonstrations in the prompts in the zero-shot scenarios.}
\label{tab:prompt}
\end{table*}

\begin{table*}[]
\centering
\resizebox{0.98\textwidth}{!}{%
\begin{tabular}{l|l}
\toprule
\textbf{Dataset} & \textbf{Finetuning data example for LLaMa2$_{sft}$} \\ \midrule
IMDB & \begin{tabular}[c]{@{}l@{}}\{"instruction": "Identify the sentiment of the following paragraph, output 'positive' or 'negative'.", \\ \\ "input": "The cast played Shakespeare.Shakespeare lost.I appreciate that this is trying to bring Shakespeare to the masses, \\ but why ruin something so good. Is it because 'The Scottish Play' is my favorite Shakespeare? I do not know. \\ What I do know is that a certain Rev Bowdler (hence bowdlerization) tried to do something similar in the \\ Victorian era.In other words, you cannot improve perfection.I have no more to write but as I have to \\ write at least ten lines of text (and English composition was never my forte I will just have to keep \\ going and say that this movie, as the saying goes, just does not cut it.", \\ \\ "output": "negative"\},\end{tabular} \\ \hline
SST-5 & \begin{tabular}[c]{@{}l@{}}\{"instruction": "Identify the sentiment of the following paragraph, \\ you have five options: 'Strong Positive', 'Positive', 'Neutral', 'Negative' or 'Strong Negative'.",\\ \\ "input": "The gorgeously elaborate continuation of The Lord of the Rings trilogy is so huge that a\\ column of words can not adequately describe co-writer/director Peter Jackson 's expanded vision of J.R.R. \\ Tolkien 's Middle-earth .", \\ \\ "output": "Strong Positive"\},\end{tabular} \\ \hline
WN18RR & \begin{tabular}[c]{@{}l@{}}\{"instruction": "Identify the semantic relation of the following word pair, \\ you have eight options: 'antonym',..., 'synonym'.", \\ \\ "input": "a.m. A.M.", \\ \\ "output": "synonym"\},\end{tabular} \\ \hline
FreeBase & \begin{tabular}[c]{@{}l@{}}\{"instruction": "Identify the semantic relation of the following word pair, \\ you have eight options: 'contains', ..., 'nutrient'.", \\ \\ "input": "Autoharp Guitar", \\ \\ "output": "group\_role"\},\end{tabular} \\ \bottomrule
\end{tabular}%
}
\caption{Finetuning data examples for LLaMa2$_{sft}$}
\label{tab:llama}
\end{table*}




\end{document}